\documentclass[journal]{IEEEtran}
\usepackage{amsmath,amsfonts}
\usepackage{array}
\usepackage[caption=false,font=normalsize,labelfont=sf,textfont=sf]{subfig}
\usepackage{textcomp}
\usepackage{stfloats}
\usepackage{url}
\usepackage{verbatim}
\usepackage{graphicx}
\usepackage{float}
\usepackage{cite}
\usepackage{makecell}
\usepackage{colortbl}
\usepackage{booktabs} 
\usepackage{threeparttable}
\usepackage[linesnumbered,ruled,vlined]{algorithm2e}
\begin{document}

\title{ToFe: Lagged Token Freezing and Reusing for Efficient Vision Transformer Inference}

\author{Haoyue Zhang, Jie Zhang, Song Guo~\IEEEmembership{Fellow,~IEEE,}
\thanks{Haoyue Zhang, Jie Zhang and Song Guo are with the Department of Computer Science and Engineering, Hong Kong University of Science and Technology, Hong Kong, SAR, China (e-mail: hzhangex@connect.ust.hk; csejzhang@ust.hk; songguo@cse.ust.hk).}
}



\maketitle

\begin{abstract}

Although vision transformers (ViT) have shown remarkable success in various vision tasks, their computationally expensive self-attention hinder their deployment on resource-constrained devices. Token reduction, which discards less important tokens during forward propagation, has been proposed to enhance the efficiency of transformer models. However, existing methods handle unimportant tokens irreversibly, preventing their reuse in subsequent blocks. Considering that transformers focus on different information among blocks, tokens reduced in early blocks might be useful later. Furthermore, to adapt transformer models for resource-constrained devices, it is crucial to strike a balance between model performance and computational overhead.
To address these challenges, in this paper, we introduce a novel Token Freezing and Reusing (ToFe) framework, where we identify important tokens at each stage and temporarily freeze the unimportant ones, allowing their lagged reusing at a later stage. Specifically, we design a prediction module for token identification and an approximate module for recovery of the frozen tokens. By jointly optimizing with the backbone through computation budget-aware end-to-end training, ToFe can adaptively process the necessary tokens at each block, thereby reducing computational cost while maintaining performance. 
Extensive experiments demonstrate that ToFe reduces the computational cost of LV-ViT model by 50\% with less than 2\% drop in Top-1 accuracy, achieving a better trade-off between performance and complexity compared to state-of-the-art methods.

\end{abstract}


\section{Introduction}
\label{intro}

\begin{figure}[t]
\centering
\includegraphics[width=0.48\textwidth]{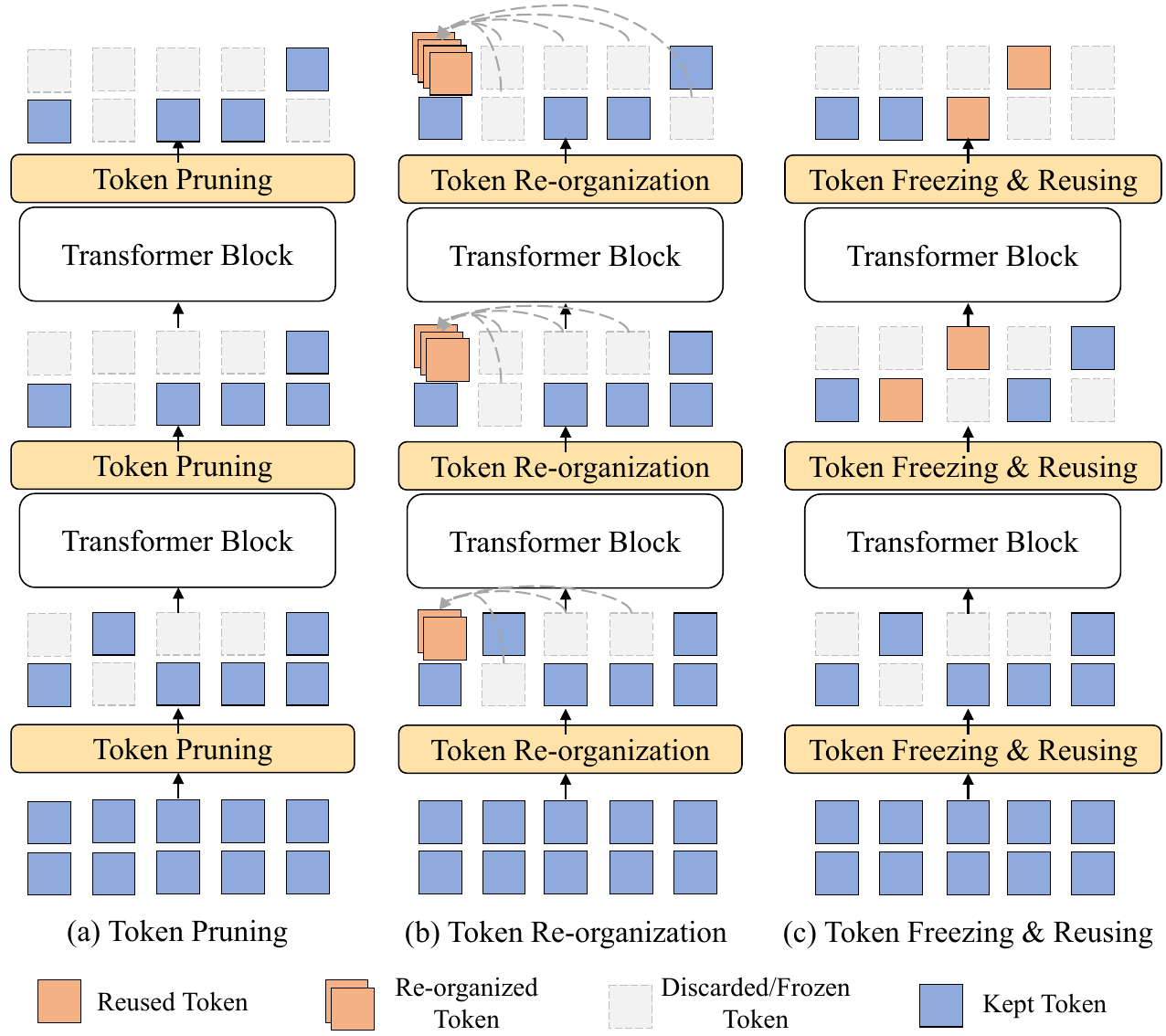}
\caption{Comparison of (a) Token pruning~\cite{rao2021dynamicvit,liu2023adaptive}, (b) Token Re-organization~\cite{bolya2022token, long2023beyond, wei2023joint}, and (c) Our token freezing and reusing. Compared with (a) and (b), our ToFe temporarily freezes ``unimportant'' tokens instead of handling the tokens directly. Some frozen tokens will be reused in later blocks if necessary, avoiding mistakenly discarding important tokens.}
\label{TokenRed}
\end{figure}

Large-scale pre-trained vision transformer (ViT) models~\cite{dosovitskiy2021an} have achieved remarkable progress in the field of vision tasks. 
However, the explosion of diverse ViT applications highlights a critical challenge: the extremely high computational cost caused by the quadratic computational complexity in the self-attention module of transformer models~\cite{bolya2022token}, which poses significant barriers to practical deployment, especially under resource-constrained circumstances.
To improve the efficiency of ViTs, numerous studies apply traditional model compression techniques like model distillation~\cite{touvron2021training,ren2023tinymim}, 
parameter pruning~\cite{ma2023llm,yu2022width} and quantization~\cite{ding2022towards}, etc., to pursue smaller sized models. 
However, traditional model compression techniques tend to prune a large portion of the model to meet tight deployment constraints, which may not guarantee optimal accuracy, especially for smaller models with long input tokens.

Unlike the above approaches that focus on building efficient transformer models, token reduction, leveraging the unique characteristics of transformer architectures and the inherent sparsity of the attention mechanism~\cite{liang2022not} to prune unimportant input tokens, has emerged as a promising approach~\cite{long2023beyond, wei2023joint,liang2022not,liu2023adaptive,zhang2024synergistic,chen2023diffrate, liusimple}. It is based on the intuition that not all tokens in the input sequences are critical for making a final prediction. Pruning these uninformative tokens within each block can increase the model's inference speed without sacrificing performance accuracy. Moreover, the removal of these informative tokens also reduces the computation and memory requirements for its subsequent blocks, leading to a linear or even quadratic reduction and bringing greater acceleration benefits. 
%
Generally, there are three key problems in implementing token reduction: 

\begin{enumerate}[]
    \item \textit{How to identify important tokens that should be processed in each transformer block, while the less important ones can be reduced? }
    \item \textit{How to handle the less important tokens (directly discard, merge, or re-organize them, etc.)? }
    \item \textit{How to find the optimal number of the reserved tokens or the reduction ratio in each block?}
\end{enumerate}
%
%

\begin{figure}[t]
\centering
\includegraphics[width=0.48\textwidth]{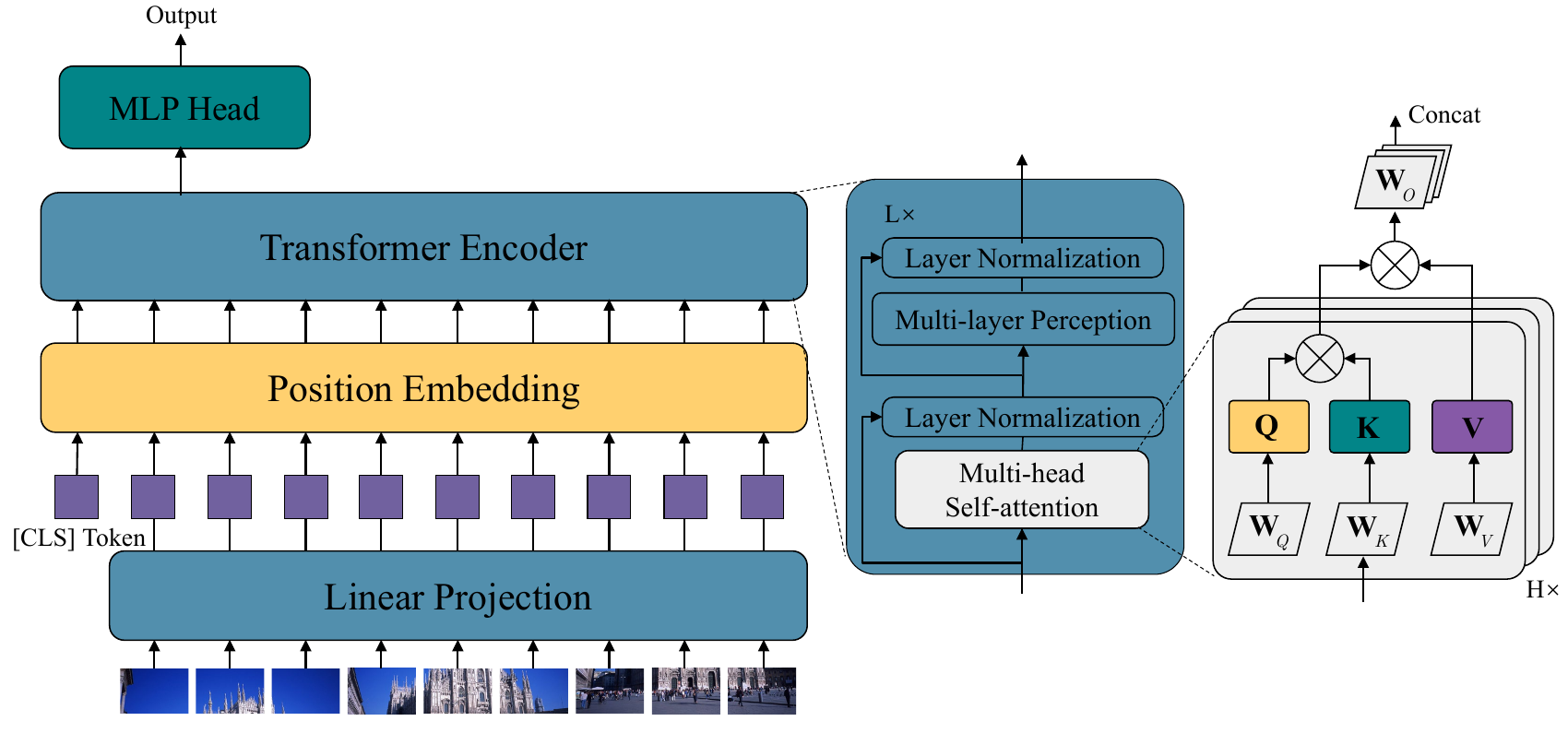}
\caption{An illustration of a typical transformer model with $L$ blocks. The input image is first split into patches, linearly projected and embedded. Then the tokens are forwarded by $L$ transformer blocks that contain layer normalization, multi-head self-attention, and multi-layer perception.}
\label{Transformer}
\end{figure}

For the first problem, since the final output of the transformer model primarily depends on the [CLS] token (i.e., Fig.~\ref{Transformer}), the task-relevant information is concentrated in the [CLS] token. Thus, the attention values of the [CLS] token to other tokens have been commonly used as a metric for evaluating token importance~\cite{liang2022not,bolya2022token, liusimple,liu2023adaptive,zhang2024synergistic,chen2023diffrate,long2023beyond, wei2023joint}. 
However, to ensure accurate predictions during inference, the [CLS] token will be forced to pay more attention to the most task-relevant tokens as the block gets deeper, meaning that using [CLS] attention values in deeper blocks is more feasible than in shallower blocks, where this characteristic may not be shown in shallower blocks. 

For the second problem, mainstream token reduction mechanisms manipulate the less important tokens in two ways: token pruning, i.e., directly discard~\cite{rao2021dynamicvit,liu2023adaptive} and token re-organization, i.e., merge~\cite{bolya2022token}, fusion~\cite{long2023beyond}, squeeze~\cite{wei2023joint}. All these methods treat the unimportant/inattentive tokens as totally useless ones and handle them irreversibly, where the reduced tokens cannot be recovered and reused in deeper blocks. 
In this case, the performance of the model would be significantly degraded due to the removal of some temporary inattentive tokens in shallower blocks.

For the last problem, most of the existing methods either require manual selection of the keeping ratio for different reduction stage~\cite{liang2022not,bolya2022token,liusimple,long2023beyond,wei2023joint}, or consider the keeping ratio as a learnable parameter~\cite{liu2023adaptive,chen2023diffrate} or optimizable variables~\cite{zhang2024synergistic}. However, due to the inaccuracy of using [CLS] token's attention values as importance proxy, both manually and automatically determined keeping ratios are usually higher in shallow blocks to avoid mistakenly pruning on useful tokens, thus hindering further acceleration of transformer models.

\begin{figure}[t]
\centering
\includegraphics[width=0.45\textwidth]{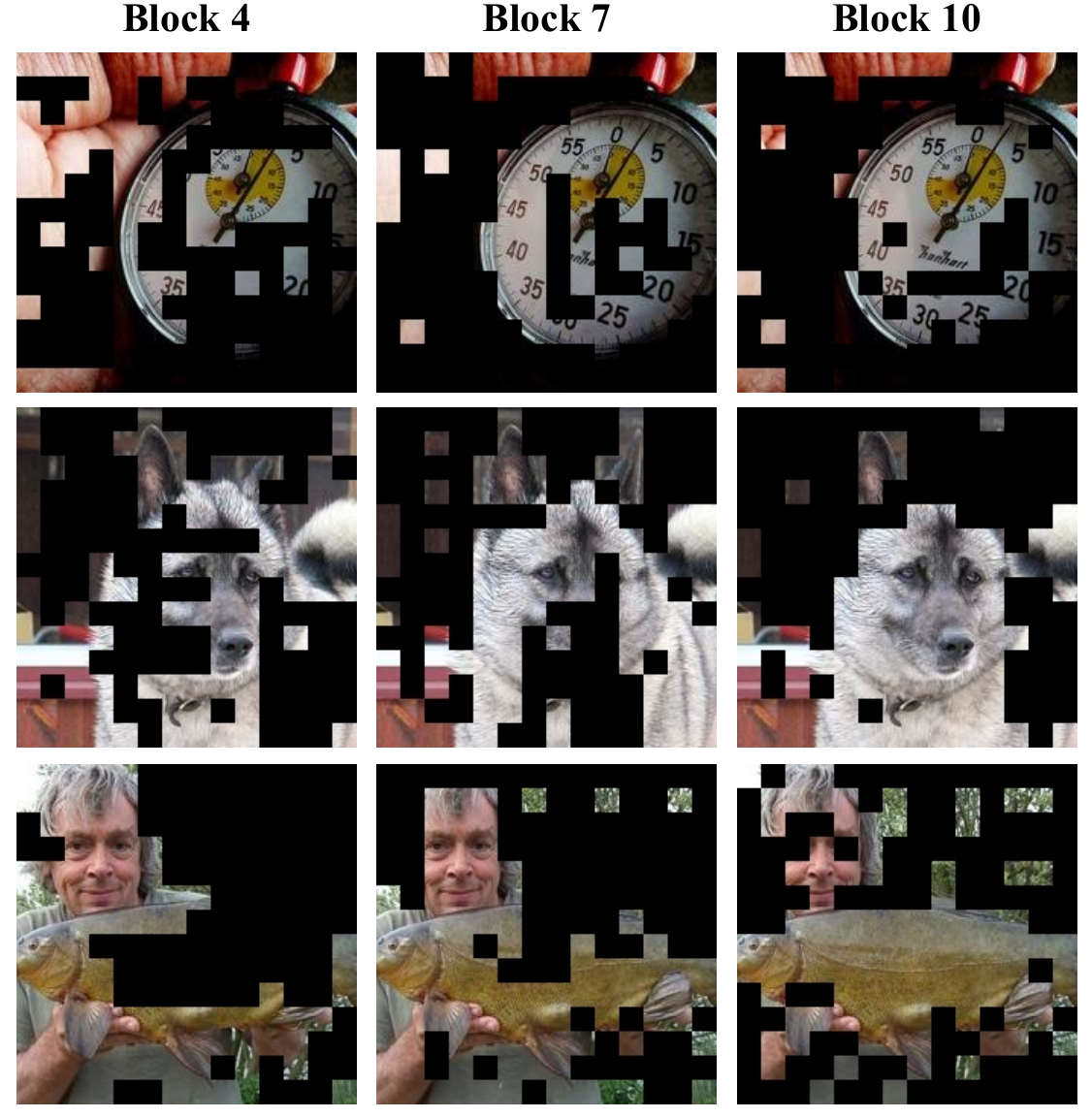}
\caption{Visualization of [CLS] Token Attention. We use DeiT-S model and three image samples from ImageNet-1K dataset. We sort the tokens based on the [CLS] attention values in the 4-th, 7-th, and 10-th blocks. In each block, tokens with top-50\% [CLS] attention value are kept, while others are removed and marked with black squares.}
\label{visattn}
\end{figure}

In fact, the attention of [CLS] token is relatively scattered at shallow blocks. For instance, we use DeiT-S model and three image samples from ImageNet-1K dataset, and simply keep the top-50\% of tokens based on the [CLS] attention values in the 4-th, 7-th, and 10-th blocks of each image, respectively. As illustrated in Fig.~\ref{visattn}, most of the informative tokens are preserved in the 10-th block (e.g., the dog's head and the entire body of the fish), but a large part of these tokens are removed in the 4-th block. Since the removed tokens cannot be used in subsequent blocks, the performance of token reduction deteriorates dramatically when we decrease the keeping ratio.

In summary, the primary limitation of existing token reduction methods is that [CLS] token's attention values cannot accurately represent the importance of tokens in shallower blocks, and the tokens reduced in shallower blocks will not be used again in deeper blocks. To tackle this problem and further solve the trade-off between model performance and model complexity, we propose lagged
\underline{\textbf{To}}ken \underline{\textbf{F}}reezing and R\underline{\textbf{e}}using (\textbf{ToFe}), a simple yet highly effective and efficient token reduction framework. Specifically, ToFe utilizes a lightweight prediction module to adaptively decide which token is important at each stage in a global computation budget-aware view, temporarily processing them only for the current stage\footnote{Similar to the previous token reduction works~\cite{liang2022not,bolya2022token,rao2021dynamicvit}, ToFe implements multi-stage token reduction. Denote that the $s$-th ($s \in {1,\cdots,S}$) token reduction operation is adopted at the $l_s$-th block, where $l_s \in \{l_1,\cdots,l_S\}$. For example, for a 16-block ViT model like LV-ViT-S~\cite{jiang2021token}, a three-stage token reduction is deployed at the 5-th, 9-th, and 13-th blocks.}. 
By temporarily freezing ``unimportant'' tokens instead of handling them directly, some frozen tokens will be reused in later blocks if necessary.
To compensate for the potential inaccuracy of the frozen tokens when skipping multiple blocks, we further introduce a lightweight approximation module to recover the error of the frozen tokens. 
With the freeze-then-reuse framework, transformer models can choose the tokens exactly needed in each block, minimizing the number of tokens calculated by transformer blocks while maintaining better performance. 

Overall, our main contributions are summarized as follows.

\begin{itemize}

\item We present ToFe, a novel Token Freezing and Reusing framework for efficient transformer inference. At each stage, all tokens are fed into a prediction module to decide which tokens to use (or reuse) in the following stage. Only the selected tokens are input to transformer blocks, while others are frozen and approximated for later blocks, thereby reducing the computational cost. 

\item We introduce a computation budget-aware training framework to jointly learn the prediction module and approximation module, enabling a globally optimized token selection phase and a flexible token recovery process.

\item We conduct extensive experiments on widely used vision transformer backbones. The experimental results show that our method reduces the computational cost of LV-ViT by 50\% and brings less than 2\% drop in Top-1 accuracy, achieving a better trade-off between model performance and model complexity than previous methods.

\end{itemize}

\section{Related Works}

\textbf{Efficient Vision Transformer Models.} Due to the high complexity of transformer models, reducing the computational cost and accelerating transformer models have become popular research topics in both academia and industry. Several works focus on efficient transformer architecture design. Specifically, DeiT~\cite{touvron2021training} introduces model distillation to compress large ViT models. 
A window-based self-attention is proposed in Swin Transformer~\cite{liu2021swin} to reduce complexity to linear with respect to the number of tokens. 
CrossViT~\cite{chen2021crossvit} proposes a dual-branch transformer architecture that processes multi-scale image patches. By combining smaller and larger patch embeddings, CrossViT achieves a balance between efficiency and accuracy.
PVT~\cite{wang2021pyramid} and CoaT~\cite{xu2021co} are both hybrid architectures that leverage convolutional operations to construct a hierarchical ViT with spatial-reduction attention. The design incorporates lightweight aggregation mechanisms that reduce computational complexity while maintaining competitive performance on vision tasks.
Unlike these architectural design approaches, ToFe focuses on token-level processing and is orthogonal in these efforts.

\textbf{Token Reduction by Token Pruning.} Given the informative redundancy of input tokens and the sparsity of the attention mechanism, unimportant tokens can be reduced during forward propagation, accelerating transformer models without additional steps. Token pruning, which identifies and directly discards unimportant tokens, is one of the key approaches in existing token reduction mechanisms. DynamicViT~\cite{rao2021dynamicvit} and IA-RED2~\cite{pan2021ia} adopt prediction modules to dynamically identify informative tokens. EViT~\cite{liang2022not} and AS-ViT~\cite{liu2023adaptive} notice that the most task-relevant information is concentrated in the [CLS] token, so the [CLS] attention is used to measure token importance. Moreover, Both METR~\cite{liusimple} and STAR~\cite{zhang2024synergistic} recognize that the attention value of the [CLS] token is not a reliable proxy for token importance, and they propose respective solutions to address this issue. Specifically, METR enhances task-related information in the [CLS] token by training early exit heads, while STAR employs a training-free approach to optimize the keep ratio at each layer, thereby preventing the discarding of important tokens.

\textbf{Token Reduction by Token Re-organizing.} The other mechanism of token reduction is token re-organization, where the unimportant tokens are compressed and integrated. This method alleviates the issue of important tokens potentially being discarded to some extent, as some of the information from the reduced tokens is merged into the retained tokens. For example, Evo-ViT~\cite{xu2022evo} and ToFu~\cite{kim2024token} merge the pruned tokens into one, while neglecting the inconsistency of these merged tokens. To address this issue, ToMe~\cite{bolya2022token} and TPS~\cite{wei2023joint} use the bipartite soft matching and squeezing method to integrate similar tokens, respectively. Moreover, PPT~\cite{wu2023ppt} and LTMP~\cite{bonnaerens2023learned} integrate pruning and merging methods together to further improve the performance of token reduction.

Unlike the aforementioned methods that handle reduced tokens in an irreversible manner, our approach uncovers the potential importance of reduced tokens in later blocks and allows for their reusing, avoiding the loss of important information, thus improving model performance. By leveraging the strengths of both token pruning and re-organization, ToFe balances computational efficiency and accuracy, providing a robust solution for accelerating transformer models. Although recent work IdleViT~\cite{xu2023no} also adopts token reusing, [CLS] attention is still utilized as importance proxy, which means it always processes self-attention for all tokens and only save MLP calculations for unimportant tokens. In contrast, ToFe predictively selects reused and frozen tokens before the transformer blocks, saving more computation by retaining the same number of tokens.

\section{Background and Motivation}

\begin{figure*}[t]
    \centering
         \subfloat[]{
		\includegraphics[width=0.4\textwidth]{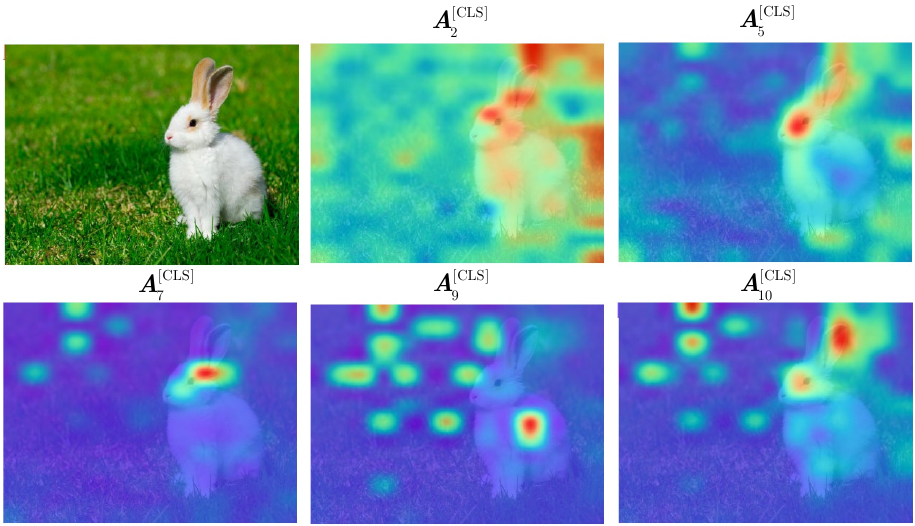}
		\label{attenmap}}
        \hspace{-0.2cm}
	\subfloat[]{
		\includegraphics[width=0.26\textwidth]{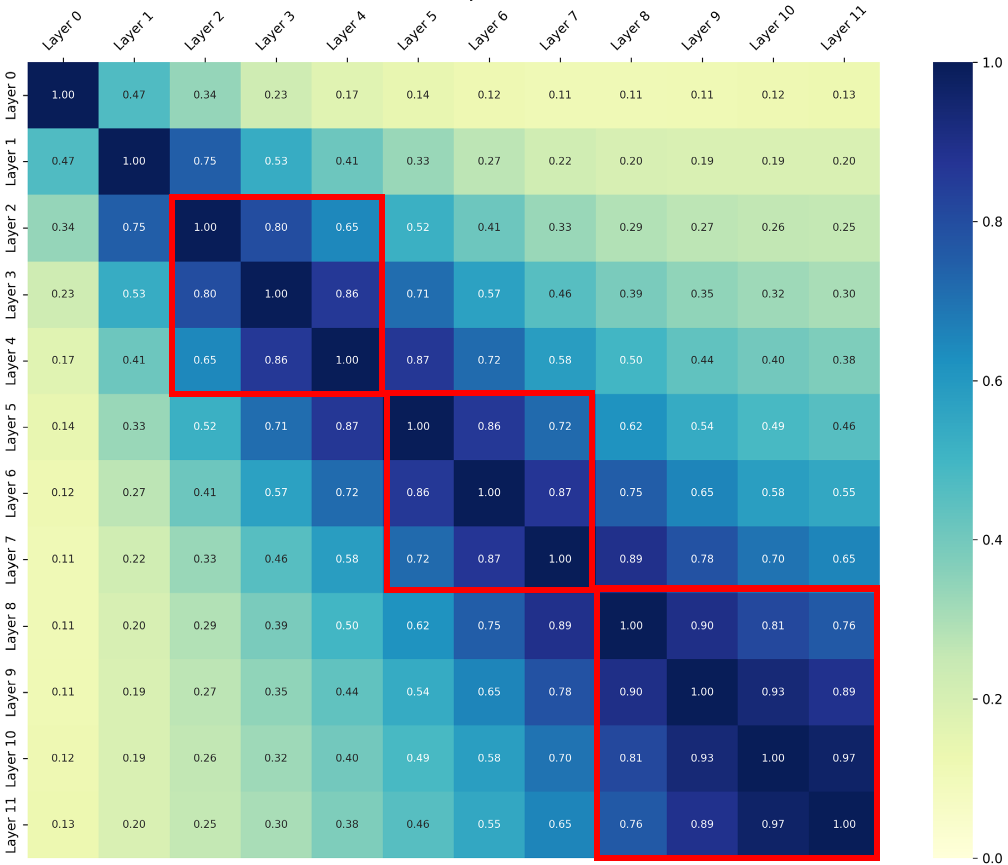}
		\label{attnsim1}}
        \hspace{-0.2cm}
        \subfloat[]{
		\includegraphics[width=0.26\textwidth]{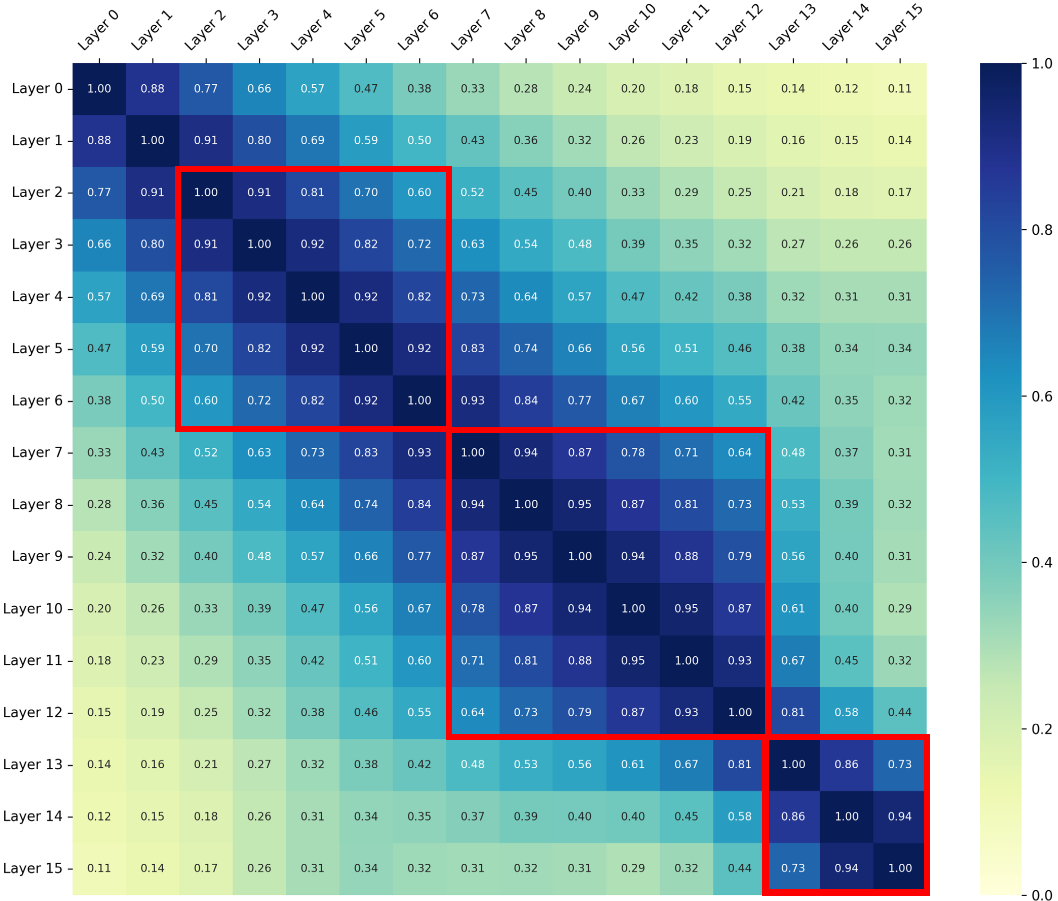}
		\label{attnsim2}}
	\caption{(a) Visualization of the [CLS] attention values in 2-th, 5-th, 7-th, 9-th and 10-th blocks of DeiT-S model. The red parts of the heat map represent areas where [CLS] attention is concentrated; (b) Similarity of token features among different blocks in DeiT-S. (c) Similarity of token features among different blocks in LV-ViT-S. The rows and columns represent the index of the block. The lighter the squares in the heat map represent the higher the similarity.}
 \label{vis}
\end{figure*}

\subsection{Preliminary: Transformer Model}

Currently, the transformer model has demonstrated outstanding performance in visual and language tasks due to its ability to capture long-term dependencies between input sequences and can easily scale its model parameters to millions or even billions~\cite{vaswani2017attention,dosovitskiy2021an,devlin2018bert}, making it one of the most popular deep learning network models.

The Vision Transformer (ViT)~\cite{dosovitskiy2021an}, in particular, suffers from significant computational costs due to high-resolution images and video data, resulting in input token numbers reaching hundreds or even thousands. On the other hand, there is more spatial and semantic redundancy in visual tokens, making research on token reduction in vision transformers more urgent. Therefore, in this paper, we present our method based on the vision transformer model.

As illustrated in Fig.~\ref{Transformer}, in a typical ViT model, an image is first split into $N$ non-overlapping patches and embedded into a $D$ dimensional feature space. Notably, an extra [CLS] token~\cite{touvron2021training} is added to the beginning of the image embedding for final classification. Then, the total input token sequence is added with a learnable position encoding and fed into the transformer encoder. A transformer encoder is stacked by $L$ blocks, each containing Layer Normalization (LN), Multi-head Self-attention (MHSA), and Multi-layer Perception (MLP)~\cite{vaswani2017attention}. The main operations inside the $l$-th block can be represented by using the following two expressions: 
\begin{equation}
{\mathbf{X}_{l-attn}} = {\mathbf{X}_l} + {\rm{MHSA}}({\rm{LN}}({\mathbf{X}_l})),
\label{eq1}
\end{equation}
\begin{equation}
{\mathbf{X}_{l + 1}} = {\mathbf{X}_{l - attn}} + {\rm{MLP}}({\rm{LN}}({\mathbf{X}_{l - attn}})),
\label{eq2}
\end{equation}
where $\mathbf{X}_l$ denotes the input tokens for the $l$-th block.

MHSA allows the ViT model to extract features from different representation spaces by using different attention heads. In each attention head, the input feature is first linearly projected into three matrices Query $\mathbf{Q}$, Key $\mathbf{K}$ and Value $\mathbf{V}$, and $\mathbf{Q}, \mathbf{K},\mathbf{V} \in \mathbb{R}^{(N + 1) \times D}$. Then, $H$ attention heads calculate the attention values parallel as follows:
\begin{equation}
{\rm{Attn}}(\mathbf{Q},\mathbf{K},\mathbf{V}) = {\rm{Softmax}}\left( {\frac{{\mathbf{Q}{\mathbf{K}^T}}}{{\sqrt {{D_h}} }}} \right)\mathbf{V},
\label{eq3}
\end{equation}
where ${D_h} = {D \mathord{\left/ {\vphantom {D H}} \right. \kern-\nulldelimiterspace} H}$ is the output dimension of a single head. Thus, due to the matrix multiplication operation in Eq.~(\ref{eq3}), the computational cost of ViT is proportional to the quadratic number of the input tokens.

The computational cost of the ViT model can be measured by Floating Point Operations (FLOPs). The FLOPs in the $l$-th Block can be represented as
\begin{equation}
{\rm{FLOPs}}_l = 4{N_l}{D^2} + 2N_l^2D + 2{N_l}D \times {D_{\rm{hidden}}},
\label{eq4}
\end{equation}
where $N_l$ is the number of the input tokens in the $l$-th Block, and the first term in Eq.~(\ref{eq4}) corresponds to the computational cost associated with the projection of $\mathbf{Q}, \mathbf{K}, \mathbf{V}$, as well as the linear projection of the attention value. The second term represents the computation of the attention value. The final term arises from the MLP operation, where ${D_{\rm{hidden}}}$ (usually $ {D_{\rm{hidden}}} = 4D$) denotes the hidden dimension in MLP layer.

In summary, a straightforward approach to reducing the computational cost of the ViT model is to reduce the number of input tokens in each block, which is the primary objective of our research.

\subsection{Motivation: Token Freezing and Reusing}
\label{motivation}

The evidence presented in Eq.~(\ref{eq4}) shows that reducing the number of tokens processed in each block save computational costs. Although we aim to reduce tokens at shallower blocks, existing methods treat reduced tokens in a irreversible manner. This raises a critical question: \textit{are the inattentive tokens reduced in the shallower blocks truly no longer useful}? 

To address this question, we visualize the [CLS] token attention values across different blocks~\cite{visualize2021}. 
As illustrated in Fig.~\ref{attenmap}, transformer model focus on different information at different blocks, e.g., the rabbit's body get less attention at the 2-th and 5-th block while is paid more attention at the 9-th block.
In this case, irreversibly reducing uninformative tokens (e.g., pruning, re-organizing) in shallower blocks would result in the loss of some informative tokens with high importance scores in deep blocks, as the reduced tokens cannot be used for subsequent blocks. For example, pruning a rabbit's body in 7-th block will render the corresponding value in 9-th block unusable. 
Worse still, if a large number of tokens are reduced, the performance of token reduction would deteriorate sharply.
Therefore, instead of treating the reduced tokens in an irreversible manner, it is preferable to freeze them temporarily and reuse them when needed.

Nevertheless, due to the fact that frozen tokens skip multiple blocks without being processed by these transformer blocks, directly reusing the frozen tokens inevitably introduces inaccuracy. This naturally leads to another problem: \textit{can we use a lightweight module, different from the transformer block, to approximate the error introduced by skipping these blocks}? In Fig.~\ref{attenmap}, preliminary observations indicate that most of the inattentive tokens change little across several blocks (e.g., from Block-5 to Block-10). Additionally, recent studies have shown that self-attention is highly correlated across adjacent transformer blocks~\cite{vig2019analyzing}. We further verify this by visualizing the similarity among tokens in Fig.~\ref{attnsim1} and \ref{attnsim2} , which is calculated by the cosine similarity among the tokens feature vectors output in each blocks. The results show high similarities among several consecutive blocks (such as the highlighted blocks by red boxes). Therefore, token values may undergo only minor changes when forwarded through several successive transformer blocks, making it feasible to approximate these changes.

\begin{figure*}[ht]
\centerline{\includegraphics[scale=0.55]{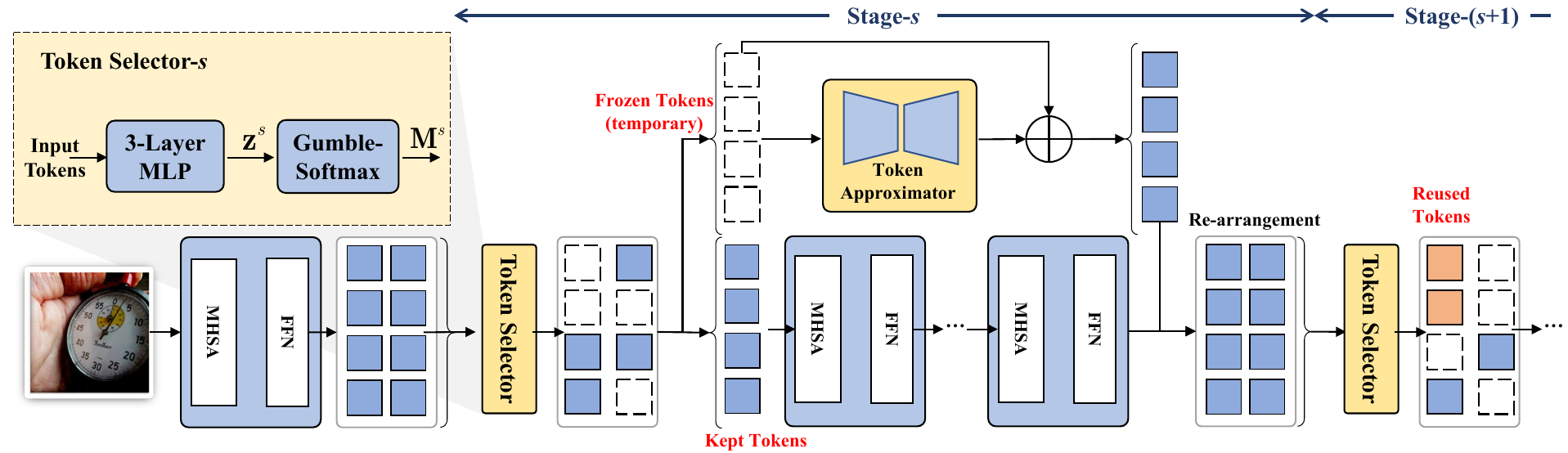}}
\caption{Overview of our proposed ToFe Framework.
The ToFe framework enhances vanilla vision transformer models (e.g., ViT~\cite{dosovitskiy2021an}, DeiT~\cite{touvron2021training}, LV-ViT~\cite{jiang2021token}) by introducing a multi-stage token freezing and reusing process. At each stage, a token selector divides tokens into kept and frozen tokens. Then, the kept tokens are processed by subsequent transformer blocks and the frozen tokens are approximated by a lightweight token approximator to compensate for the error caused by skipping those blocks.}
\label{ToFe}
\end{figure*}

\section{Token Freezing and Reusing for Efficient Transformer}

Based on the above analysis, we propose a novel inference acceleration framework, ToFe, for 
adaptively selecting important tokens in the current stage while freezing others and reusing them when needed, thus achieving a better balance between model performance and complexity.

\subsection{Accelerating Transformer with Token Freezing and Reusing}\label{AA}

We first introduce the proposed ToFe framework. As shown in Fig.~\ref{ToFe}, ToFe is implemented based on a vanilla vision transformer (e.g., ViT~\cite{dosovitskiy2021an}, DeiT~\cite{touvron2021training}, LV-ViT~\cite{jiang2021token}, etc.) as the backbone. For a $S$-stages token reduction process, we use $S$ token selectors and $S-1$ token approximators. Denote that $s$-th ($s \in \{1,\cdots,S\}$) token reduction operation is adopted before $l_s$-th block, where $l_s \in \{l_1,l_2,\cdots,l_S\}$. At stage-$s$ $(s<S)$, two main modules: \textit{token selector} and \textit{token approximator}, are deployed, while the last stage only deploys one token selector. 

Note that the [CLS] token is always kept during the forward propagation, the token selector is responsible for dividing the $N$ patch tokens into kept tokens and frozen tokens, i.e., taking the patch tokens ${\mathbf{X}}\in \mathbb{R}^{N \times D}$ as input and divide them into the kept tokens $\mathbf{X}_{\texttt{keep}}$ and the frozen tokens $\mathbf{X}_{\texttt{freeze}}$. The kept patch tokens are first concatenated with the [CLS] token, i.e., $\mathbf{X}_{\texttt{keep}} \leftarrow [\mathbf{X}_{\texttt{CLS}},\mathbf{X}_{\texttt{keep}}]$. Then $\mathbf{X}_{\texttt{keep}}$ will be forwarded to the subsequent transformer blocks, while the frozen tokens $\mathbf{X}_{\texttt{freeze}}$ directly skip the following blocks and are fed into a lightweight token approximator to compensate for the error caused by skipping those blocks. Note that we do not deploy token approximator at stage-$S$. At the stage-$s$ $(s<S)$, the token selector $s$ divides tokens before $l_s$-th block, then the overall operations of ToFe are given by:
\begin{equation}
\mathbf{X} \leftarrow {\rm{ViTBlock}}_{l}(\mathbf{X}_{\texttt{keep}}), l_s \le l < l_{s+1},
\label{eq5}
\end{equation}
\begin{equation}
\mathbf{X}_{\texttt{approx}} \leftarrow (\mathbf{X}_{\texttt{freeze}} + {\rm{Approx}}_{s}(\mathbf{X}_{\texttt{freeze}})),
\label{eq6}
\end{equation}
\begin{equation}
\mathbf{X} \leftarrow {\rm{Rearrange}}(\mathbf{X},\mathbf{X}_{\texttt{approx}}),
\label{eq7}
\end{equation}
where ${\rm{ViTBlock}}_{l}$ represents the $l$-th ViT Block of the backbone model, ${\rm{Approx}}_{s}$ is the $s$-th token approximator, and $\mathbf{X}_{\texttt{approx}}$ is the output of the token approximator, which means the approximate residual of the frozen tokens. Since dividing tokens into $\mathbf{X}_{\texttt{freeze}}$ and $\mathbf{X}_{\texttt{keep}}$ break the relative spacial relationship among tokens, we need ${\rm{Rearrange}}()$ to recover the arrangement of $\mathbf{X}$ and $\mathbf{X}_{\texttt{freeze}}$. Then, instead of processing all tokens in subsequent ViT blocks, only the kept tokens other than the frozen tokens will be computed to accelerate the inference process.

\subsection{Token Freezing and Reusing with Token Selector and Token Approximator}
\label{module}

The most important characteristic of ToFe is its ability to adaptively identify the temporarily useful tokens in the current stage. To achieve this, we design a lightweight prediction module to generate a decision mask for each patch token. We simply employ a Multilayer Perceptron (MLP) followed by a Gumbel Softmax~\cite{jang2016categorical,maddison2016concrete} to make the binary decision mask differentiable:
\begin{equation}
\mathbf{z}^s = {\rm{Softmax}}({\rm{MLP}}(\mathbf{X})) \in \mathbb{R}^{N \times 2},
\label{eq8}
\end{equation}
\begin{equation}
\mathbf{M}^{s} = {\rm{GumbleSoftmax}}(\mathbf{z^s}) \{0,1\}^{N},
\label{eq9}
\end{equation}
where $\mathbf{z}^s$ is a score vector to indicate the importance of patch tokens at stage-$s$. $\mathbf{M}^{s} \in \{0,1\}^{N}$ is a set of binary decision masks to indicate whether to keep or freeze each patch token.

Since the frozen tokens introduce additional inaccuracy in the forward propagation, they may lead to a performance degradation. Therefore, an approximate module is necessary to bridge the gap caused by freezing temporary inattentive tokens. However, this module should be simple and lightweight to avoid excessive computational costs. 
To quantify the approximation gap, we define $\Delta \mathbf{X} = {\mathbf{X}_{l + 1}} - {\mathbf{X}_l}$ as the variation of token features between two consecutive transformer blocks, and further introduce $\varepsilon  = {{{{\left\| {\Delta \mathbf{X}} \right\|}_2}} \mathord{\left/
 {\vphantom {{{{\left\| {\Delta X} \right\|}_2}} {{{\left\| {{\mathbf{X}_l}} \right\|}_2}}}} \right.
 \kern-\nulldelimiterspace} {{{\left\| {{X_l}} \right\|}_2}}}$ as the relative change in token features. According to the Universal Approximation Theorem~\cite{cybenko1989approximation}, MLPs with sufficient complexity (adequate hidden layers/neurons) can approximate any continuous function. Here, we model frozen tokens updates as function $\Delta X = f({X_l})$. As observed in Sec.~\ref{motivation}, most of the temporary inattentive tokens undergo minimal changes $\varepsilon$ across several consecutive blocks. When updates exhibit a small $\varepsilon$, $f$ likely resembles linear or simple nonlinear transformations within the input space. MLPs prove particularly effective for such approximations, as minor updates seldom involve complex higher-order nonlinearities. The above analysis inspired us to use a simple but efficient residual architecture to approximate the frozen tokens, where a bottleneck structure module with a 2-layer MLP (Token Approximator in Fig.~\ref{ToFe}) is used to approximate the sparse variation of the frozen tokens. 


\subsection{Training Objective}
In this part, we introduce the training objective to ensure that the parameters of token selector and token approximator can be well learned.

First, to minimize the impact of reusing frozen tokens, we use the original backbone ViT model as a teacher model and align the output of the token approximator as closely as possible to the teacher model's output. Specifically,  we design an approximation loss to minimize the discrepancy between the approximated frozen tokens and those output by the teacher model, which can be viewed as a kind of self-distillation loss~\cite{jiang2021all}. Given one mini-batch $B$ of the training samples, the approximate loss is written as:
\begin{equation}
        {\cal L}_{{\rm{apr}}} = \frac{1}{{\left| B \right|}}\sum\limits_{b = 1}^{\left| B \right|} \sum\limits_{s = 1}^S {\frac{1}{{{N^s_b}}}\left[ {(\mathbf{1}-{\mathbf{M}^s_b}) \odot {{({\mathbf{x}^s_b}^\prime  - {\mathbf{x}^s_b})}}} \right]^2},
    \label{eq10}
\end{equation}
where ${N^s_b} = \sum\nolimits_{i = 1}^N {\mathbf{M}_{i,b}^s}$ is the total number of the frozen tokens for the $b$-th sample at stage-$s$, $\odot$ is the Hadamard product, ${\mathbf{X}^s}^\prime$ and ${\mathbf{X}^s}$ are the output token after the stage-$s$ of ToFe and the teacher model, respectively. Then, we adopt the standard cross-entropy loss to optimize the whole framework:
\begin{equation}
{{\cal L}_{{\rm{cls}}}} = \frac{1}{{\left| B \right|}}\sum\limits_{b = 1}^{\left| B \right|}{\rm{CrossEntropy}}({\mathbf{y}_b}, {\bar{\mathbf{y}}}_b),
\label{eq11}
\end{equation}
where $\mathbf{y}_b$ and ${\bar{\mathbf{y}}}_b$ are the final prediction of ToFe and the ground truth, respectively.

Finally, the overall training objective is the combination of the above loss functions:
\begin{equation}
    {{\cal L}} = {\lambda _{{\rm{cls}}}}{{\cal L}_{{\rm{cls}}}}{\rm{ + }}{\lambda _{{\rm{apr}}}}{{\cal L}_{{\rm{apr}}}},
\label{eq12}
\end{equation}
where ${\lambda _{{\rm{cls}}}}$ and ${\lambda _{{\rm{apr}}}}$ are hyper-parameters. Note that ${\cal L}:={{\cal L}({\mathbf{y}_b},f({\mathbf{x}_b};\mathbf{w},\mathbf{M}))}$, where $f(\cdot)$ represents the ViT model with ToFe, $\mathbf{w}$ and $\mathbf{M}$ are parameters of backbone ViT model and decision masks being optimized.

\section{Computation Budget-aware End-to-end Optimization}
\label{optimize}

In realistic applications, it is usually required for transformer models to be able to efficiently adapt to various resource-constraint edge devices, i.e., with limited computation budgets~\cite{zhang2021edge}. 
Therefore, we improve ToFe to optimally balance token freezing and reusing at each stage while adhering to a target computational cost, thereby automatically optimizing the trade-off between accuracy and complexity within a given computation budget. 
Specifically, we introduce an additional computation budget constraint in original training objective of ToFe to ensure a more flexible control of the token reduction ratio.
%
We provide two versions for computation budget-aware inference: \texttt{instance-adaptive ToFe} for sigle-instance inference and \texttt{batch-adaptive ToFe} for batch parallel inference, respectively.

Given one mini-batch $B$ of the training samples, the whole optimization problem can be changed to minimize the training objective Eq.~(\ref{eq12}) under a computation budget constraint, which is given by:
\begin{equation}
\begin{aligned} 
&{\mathop {\min }\limits_{\mathbf{w},\mathbf{M}}} \quad {{\cal L}({\mathbf{y}_b},f({\mathbf{x}_b};\mathbf{w},\mathbf{M}))}\\
&\begin{array}{r@{\quad}r@{}l@{\quad}l}
s.t. &\frac{1}{{\left| B \right|}}\sum\limits_{b = 1}^{\left| B \right|} {\sum\limits_{l = 1}^L {{\cal F}({N_{l,b}})}  \le \rm{targetFLOPs}} \\
\end{array}
\end{aligned}
\label{eq13}
\end{equation}
where the constraint guarantees that the computational cost is within a given computation budget $\rm{targetFLOPs}$. The ${\cal F}({N_{l,b}})$ is a function to calculate the actual computational cost of one block $l$ for the $b$-th sample given by Eq.~(\ref{eq4}), where ${N_{l,b}} = \sum\nolimits_{i = 1}^N {\mathbf{M}_{i,b}^s}$, $l_s \le l < l_{s+1}$. ${\mathbf{M}_{i,b}^s}$ is the mask for the $b$-th sample's $i$-th token at stage-$s$.

Since the goal of ToFe is to optimize the parameters of the token predictor modules during training, we transfer the computation budget constraint into one part of the training objective. Then we introduce a computation budget-aware loss to measure the gap between the actual computational cost and the target budget:
\begin{equation}
{{\cal L}_{{\rm{FLOPs}}}} = \frac{1}{{\left| B \right|}}\sum\limits_{b = 1}^{\left| B \right|}{\left[ \sum\limits_{l = 1}^L {{\cal F}({N_{l,b}})} - {\rm{targetFLOPs}}\right]^2}.
\label{eq14}
\end{equation}

Then we combine the computation budget-aware loss with Eq.~(\ref{eq12}) to form the overall training objective:
\begin{equation}
{\cal L_{\rm{total}}} = {\lambda _{{\rm{cls}}}}{{\cal L}_{{\rm{cls}}}}{\rm{ + }}{\lambda _{{\rm{apr}}}}{{\cal L}_{{\rm{apr}}}}{\rm{ + }}{\lambda _{{\rm{FLOPs}}}}{{\cal L}_{{\rm{FLOPs}}}},
\label{eq15}
\end{equation}
where ${\lambda _{{\rm{FLOPs}}}}$ is a hyper-parameter to balance the model accuracy and complexity.

\textbf{Training.} Although we have a well-defined model architecture and training objective, it is still non-trivial to implement in practice during training because the operation of dividing tokens is indifferentiable. Furthermore, since the generated binary decision mask is various corresponding to different input samples, simply using the tokens where ${\mathbf{M}_{i}^s = 1}$ would result in a non-uniform number of tokens within one training batch, which prevents parallel training. Thus, we first convert the overall operations of ToFe into a differentiable form:
\begin{equation}
\mathbf{X} \leftarrow \mathbf{M}^{s} \odot \rm{ViTBlocks}(\mathbf{X},\mathbf{M}^{s}) + (1 - \mathbf{M}^{s}) \odot (\mathbf{X} + \Delta \mathbf{X}),
\label{eq16}
\end{equation}
where  $\Delta \mathbf{X} = {\rm{Approx}}_{s}(\mathbf{X})$. We change token freezing into token masking~\cite{rao2021dynamicvit} with ${\mathbf{M}_{i}}$ to preserve the gradient back-propagation chain and prevent the interaction between all frozen tokens and the kept tokens. To do so, we modify the calculation of the attention value in the Softmax operation of self-attention module(i.e., Eq.~(\ref{eq3})) as follows:
\begin{equation}
{\mathbf{G}_{i,j}} = \begin{cases}
1, & i = j,\\
\mathbf{M}_j^s, & i \ne j.
\end{cases}
\label{eq17}
\end{equation}
\begin{equation}
\mathbf{S} = {\frac{{\mathbf{Q}{\mathbf{K}^T}}}{{\sqrt {{D_h}} }}},{\rm{Attn}}{(\mathbf{Q,K,V})_{i,j}} = \frac{{\exp ({\mathbf{S}_{i,j}}){\mathbf{G}_{i,j}}}}{{\sum\nolimits_{k = 1}^N {\exp ({\mathbf{S}_{i,k}}){\mathbf{G}_{i,k}}} }}\mathbf{V},
\label{eq18}
\end{equation}
where $\mathbf{S}$ is the original attention map before Softmax, $\mathbf{G}_{i,j}$ is the constructed attention mask. $\mathbf{G}_{i,j} = 1$ means the $j$-th token will contribute to the update of the $i$-th token, which prevents the interaction between all frozen tokens and the other tokens. Therefore, Eqs.~(\ref{eq17}),~(\ref{eq18}) maintain the back propagation of the loss function's gradient and allow parallel training.

\textbf{Inference.} The well-trained token selector can adaptively generate favorable freezing and reusing decisions for any input instances, which is suitable for single-instance input scenarios. However, since the binary decision mask varies according to different input instances, the location of frozen tokens and kept tokens varies in different input instances. Thus, instance-adaptive ToFe cannot be adopted for parallel inference. To adapt ToFe to single-instance and parallel inference scenarios, we provide two versions of computation budget-aware ToFe: \texttt{instance-adaptive ToFe} and \texttt{batch-adaptive ToFe}. The detailed process is summarized in Algorithm~\ref{algo1}. 

For \texttt{instance-adaptive ToFe}, given an input instance to do inference, we can use the binary decision mask $\mathbf{M}^{s}$ output by the token predictor at each stage to divide frozen tokens and kept tokens. $\mathbf{M}_{i}^{s} = 1$ means that $i$-th token should be calculated through transformer blocks at stage $s$, while $\mathbf{M}_{i}^{s} = 0$ means $i$-th token should be frozen and fed to the token approximator. 

The \texttt{batch-adaptive ToFe} is designed for parallel inference. At stage-$s$, We denote $N^s$ as the average number of tokens used as measured during training, which can be calculated by ${N^s = \frac{1}{{\left| B \right|}}\sum\nolimits_{b = 1}^{\left| B \right|}\sum\nolimits_{i = 1}^N {\mathbf{M}_{i,b}}^s}$. The parallel inference utilizes the recorded average number of kept tokens and the importance scores $\mathbf{z}^s$ obtained by the token predictor in Eq.~(\ref{eq8}). Then, we sort the tokens by the importance scores $\mathbf{z}^s$, keeping Top-$N^s$ tokens and freezing the others.

\begin{algorithm}[t] 
\caption{Computation Budget-aware Inference}
\label{algo1}
  \KwIn{Input instance(s) $\mathbf{X}$, Batch size $B$, Token reduction location $\{l_1,l_2,\cdots,l_S\}$, The average number of tokens $N^s$ measured during training;}
  \KwOut{Final Prediction $\mathbf{y}$;}

    \For {$l=1,2,\cdots,L$}
    {
        \If{$l \in \{l_1,l_2,\cdots,l_S\}$} 
        {
        \tcp{Token freezing and reusing}
            \If(\tcp*[h]{Not the last stage, recover the frozen token}){$l<l_S$}  
            {
                Find stage-$s$ where the block-$l$ is located\;
                Recover the frozen tokens using 
                Eq.~(\ref{eq6})\;
                Rearrange tokens using Eq.~(\ref{eq7})\;
            }
            \If(\tcp*[h]{instance-adaptive}){$B == 1$}     
            {
                Obtain $\mathbf{M}^s$ according to Eq.~(\ref{eq8}), Eq.~(\ref{eq9})\;
                Divide the tokens into $\mathbf{x}_{\texttt{keep}}$ and $\mathbf{x}_{\texttt{freeze}}$;
            }
            \Else(\tcp*[h]{batch-adaptive})   
            {
                Obtain $\mathbf{z}^s$ according to Eq.~(\ref{eq8})\;
                Sort tokens according to $\mathbf{z}^s$\;
                Select top-$N^s$ as the kept tokens $\mathbf{x}_{\texttt{keep}}$;
            }
            $\mathbf{x} \leftarrow \text{ViTBlocks}_l(\mathbf{x}_{\texttt{keep}})$\;
        }
        \Else
        {
            $\mathbf{x} \leftarrow \text{ViTBlocks}_l(\mathbf{x})$\;
        }
    }
\end{algorithm}

\section{Experiment}

\begin{table}[t]
\centering
\caption{Comparisons with SOTA token reduction methods.}
\label{mainres}
\begin{threeparttable}
    \begin{tabular}{lcccc}
    \toprule
        Model & \makecell{GFLOPs} & \makecell{Throughput\\(imgs/s)} & \makecell{Top-1 \\ Acc.(\%)} & \makecell{Acc. ↓ \\ (\%)} \\
        \midrule
        \textbf{LV-ViT-S} & 6.6   & 1718.6  & 83.3  & - \\
        \midrule
        EViT~\cite{liang2022not} & 3.4 & 3157.3 & 75.1  &  -8.2 \\
        DynamicViT~\cite{rao2021dynamicvit}  & 3.4  & 3348.0  & 77.9  & -5.4  \\
        AS-ViT~\cite{liu2023adaptive} & 3.4   & 3087.0 & 63.5  & -19.8  \\
        IdleViT~\cite{xu2023no} & 3.3 & 3365.7 & 79.5 & -3.8 \\
        \rowcolor[rgb]{ .749,  .749,  .749} ToFe (Ours) & 3.3  & 3265.1 & \textbf{81.9} & \textbf{-1.4 } \\
        \midrule
        \midrule
        \textbf{LV-ViT-M} & 12.7  & 991.0 & 84.1  & - \\
        \midrule
        EViT~\cite{liang2022not}  & 6.1   & 1937.4 & 78.4 & -5.7  \\
        DynamicViT~\cite{rao2021dynamicvit} & 6.1   & 2124.9 & 78.0  & -6.1  \\
        AS-ViT~\cite{liu2023adaptive} & 6.1   & 2119.9 & 80.2 & -3.9  \\\
        IdleViT~\cite{xu2023no} & 6.1 & 2013.8 & 81.9 & -2.2 \\
        \rowcolor[rgb]{ .749,  .749,  .749} ToFe (Ours) &   6.1 & 1964.2  &   \textbf{82.8}  & \textbf{-1.3} \\
        \midrule
        \midrule
        \textbf{DeiT-S} & 4.6   & 2383.9  & 79.9  & - \\
        \midrule
        EViT~\cite{liang2022not}  & 2.0   & 5442.1 & 68.6  & -11.3  \\
        DynamicViT~\cite{rao2021dynamicvit} & 2.1   & 5607.9 & 75.2  & -4.6  \\
        AS-ViT~\cite{liu2023adaptive} & 2.1   & 5411.3 & 75.6  & -4.3  \\
        ToMe~\cite{bolya2022token}  & 2.0   & 5610.1 & 76.5  & -3.3  \\
        PPT~\cite{wu2023ppt}	& 2.3	& 4269.9	& 77.0	& -2.9   \\
        LTMP~\cite{bonnaerens2023learned}	& 2.0	& 5598.4	& 76.2	& -3.7 \\
        IdleViT~\cite{xu2023no} & 2.0	&  5603.7	& 77.6	& -2.3 \\
        \rowcolor[rgb]{ .788,  .788,  .788} ToFe (Ours) & 2.0  & 5477.3 & \textbf{77.7} & \textbf{-2.2 } \\
        \midrule
        \midrule
        \textbf{DeiT-B} & 17.6  & 810.2 & 81.9  & - \\
        \midrule
        EViT~\cite{liang2022not}  & 8.7   & 1646.1 & 74.6  & -7.3  \\
        DynamicViT~\cite{rao2021dynamicvit} & 8.6   & 1942.8 & 77.5  & -4.3  \\
        AS-ViT~\cite{liu2023adaptive} & 8.7   & 1685.4 & 78.7  & -3.1  \\
        ToMe~\cite{bolya2022token}  & 8.7   & 1876.2 & 77.9  & -3.9  \\
        PPT~\cite{wu2023ppt}	& 8.7	& 1458.6	& 68.6	& -13.3 \\
        LTMP~\cite{bonnaerens2023learned}	& 8.6	& 1865.7	& 75.9	& -6.0 \\
        IdleViT~\cite{xu2023no}	& 8.6	& 1895.2	& 79.4	& -2.5 \\
        \rowcolor[rgb]{ .788,  .788,  .788} ToFe (Ours) & 8.6  & 1689.1 & \textbf{79.6} & \textbf{-2.3 } \\
    \bottomrule
    \end{tabular}
    \begin{tablenotes}    
        \footnotesize               
        \item[1] We set different configurable hyperparameters of each method to get similar computation cost. Specifically, we set different token keeping ratio $r$ for EViT~\cite{liang2022not}, DynamicViT~\cite{rao2021dynamicvit}, ToMe~\cite{bolya2022token}, IdleViT~\cite{xu2023no}, and PPT~\cite{wu2023ppt}, while setting different target computation budget for AS-ViT~\cite{liu2023adaptive}, LTMP~\cite{bonnaerens2023learned} and our ToFe.
      \end{tablenotes}            
    \end{threeparttable}       
\end{table}

\subsection{Experimental Setup.}

\textbf{Implementation Details} The experiments are conducted using two popular ViT backbones, DeiT~\cite{touvron2021training} and LV-ViT~\cite{jiang2021token}, both pre-trained on ImageNet-1K dataset~\cite{deng2009imagenet}. We employ a three-stage token reduction strategy analogous to existing methods. We employ a three-stage token reduction strategy analogous to existing methods. Specifically, for the DeiT models, we insert the token selector before the 4-th, 7-th, and 10-th blocks, and the token approximator at the 7-th and 10-th blocks. We use the pre-trained models to initialize the backbone models and jointly train the entire model with the loss function $\cal L_{\rm{total}}$ as defined in Eq.~(\ref{eq15}) for 30 epochs. The hyper-parameters are set as follows: ${\lambda _{{\rm{cls}}}} = 1.0 $, ${\lambda _{{\rm{apr}}}} = 2.0$, and ${\lambda _{{\rm{FLOPs}}}} = 5.0$. The initial learning rate is set to 1e-4 and decayed to 2e-6 using a cosine annealing schedule. Additionally, the parameters of the backbone models were fixed during the first 5 epochs.

\textbf{Evaluation Metrics.} To evaluate the trade-off between model accuracy and complexity for a given computation budget, we choose Giga FLOPs (GFLOPs) and inference throughput as the primary evaluation metrics for model complexity. Inference throughput is measured on a single NVIDIA RTX4090 GPU with a fixed batch size of 128. The throughput is averaged over 100 inferences using randomly selected images from the ImageNet-1K dataset.

\textbf{SOTA Token Reduction Methods and Backbone Models.} we compare our proposed ToFe with the state-of-the-art token reduction methods, including pruning-based methods (i.e., EViT~\cite{liang2022not}, DynamicViT~\cite{rao2021dynamicvit}, AS-ViT~\cite{liu2023adaptive}), merging-based methods (i.e., ToMe~\cite{bolya2022token}, PPT~\cite{wu2023ppt}, LTMP~\cite{bonnaerens2023learned}), and one reusing-based method, IdleViT~\cite{xu2023no}. Meanwhile, we further compare the model complexity and accuracy of ToFe with other SOTA backbone models, including CoaT~\cite{xu2021co}, CrossViT~\cite{chen2021crossvit}, PVT~\cite{wang2021pyramid}, SwinTransformer~\cite{liu2021swin}, T2T-ViT~\cite{yuan2021tokens}, LV-ViT~\cite{jiang2021token}, CPVT~\cite{chu2021conditional}, RegNetY~\cite{radosavovic2020designing}, EfficientNet~\cite{tan2019efficientnet}, and NFNet~\cite{brock2021high}. 

\subsection{Main Results}

\textbf{Comparison with SOTA Token Reduction Methods.} The main results are summarized in Table~\ref{mainres}, where we compare ToFe with SOTA token reduction methods. We test the Top-1 accuracy (Acc.), FLOPs, and throughput across four different backbone models: DeiT-S, DeiT-B, LVViT-S, LVViT-M. For a fair comparison, different compression ratios for different methods are set to achieve approximately 100\% acceleration of the backbone models, then the Top-1 accuracy is compared. Table~\ref{mainres} shows that ToFe consistently outperforms the other token reduction methods. Specifically, for a 50\% reduction in the computational cost of the backbone models, ToFe achieves an approximate or less than 2\% drop in Top-1 accuracy for the DeiT or LV-ViT, respectively. In contrast, other methods show accuracy drops ranging from 3.1\% to 19.8\%. These results demonstrate the superior performance of ToFe in accelerating transformer models while maintaining high accuracy.

\begin{figure}[t]
    \centering
        \subfloat[]{
		\includegraphics[width=1.5in]{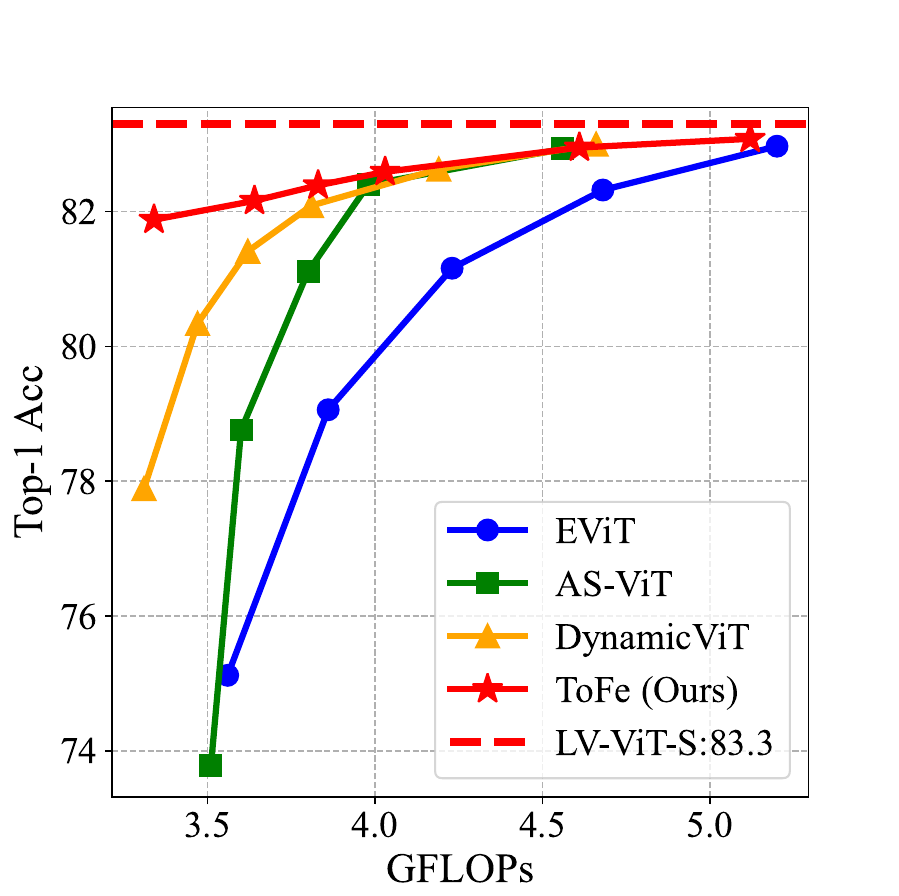}
		\label{LVVIT-S}}
	\subfloat[]{
		\includegraphics[width=1.5in]{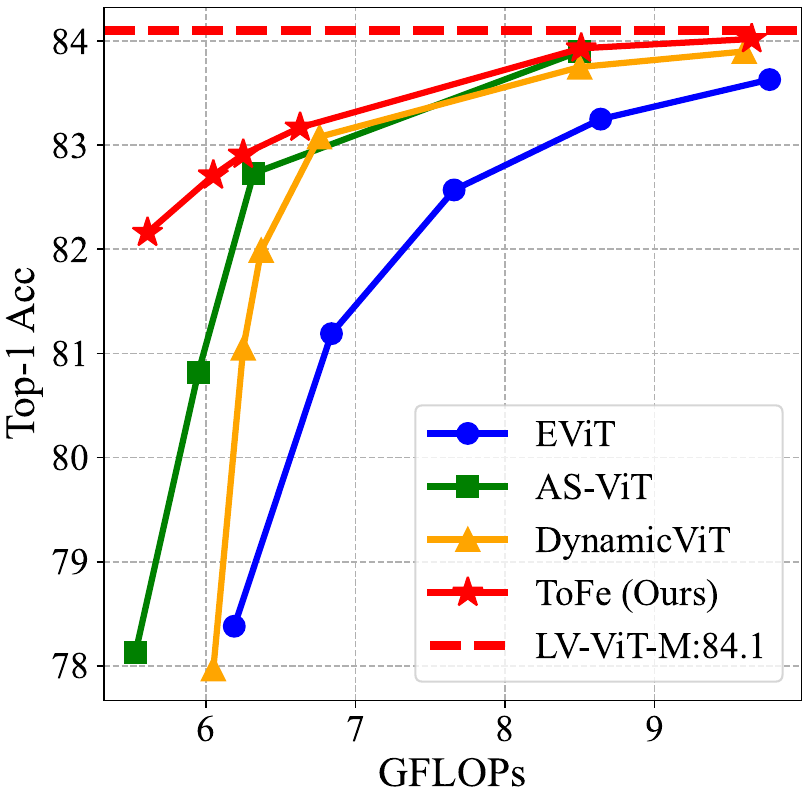}
		\label{LVVIT-M}}
	\caption{Trade-off between FLOPs and Top-1 accuracy of different token reduction methods on (a) LV-ViT-S; (b) LV-ViT-M.}
 \label{FLOPSvsACC}
\end{figure}

\begin{figure}[t]
\centering
\includegraphics[width=0.45\textwidth]{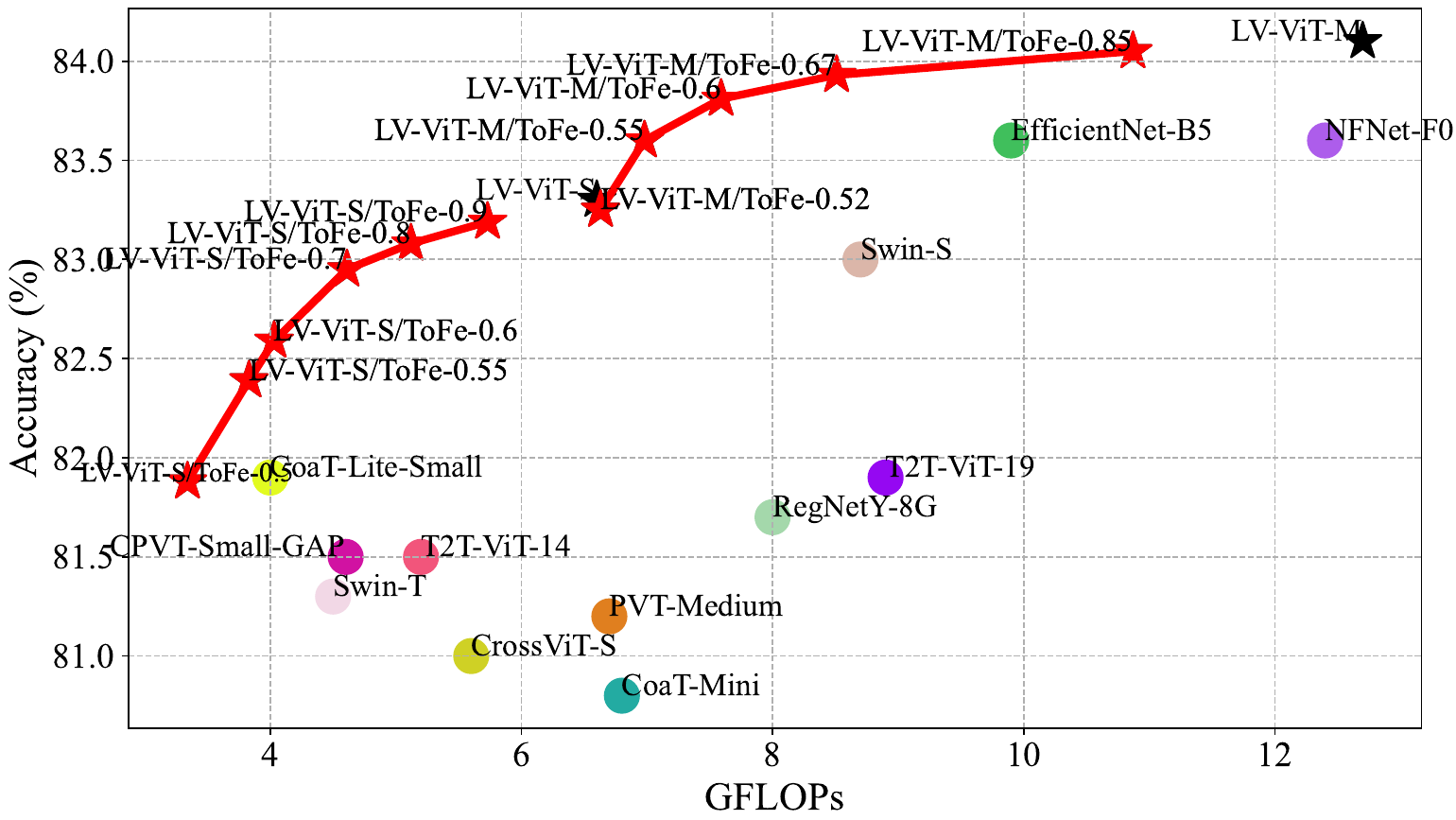}
\caption{Comparison of trade-offs between model complexity and accuracy for different models.}
\label{FLOPSvsAccMODELS}
\end{figure}

\textbf{Trade-off between the Model Complexity and Accuracy.} To further demonstrate the advantages of ToFe regarding the trade-off between model complexity and accuracy, we set different keep ratios $r$ for EViT~\cite{liang2022not}, DynamicViT~\cite{rao2021dynamicvit} and different target computation budgets for AS-ViT~\cite{liu2023adaptive} and our proposed ToFe. Fig.~\ref{FLOPSvsACC} illustrates the FLOPs and Top-1 accuracy of different token reduction methods under different computation budget settings. For both LV-ViT-S and LV-ViT-M models, our method shows a slight decrease in Top-1 accuracy as the computation budget is reduced, while other methods show a significant decrease in accuracy. Notably, when FLOPs are reduced to approximately half of the backbone models (e.g., 3.3 GFLOPs for LV-ViT-S), existing methods experience a steep decline in accuracy. By simply adjusting the computational budget according to the computational power of different devices, ToFe not only automatically achieves a superior complexity-accuracy trade-off, but also avoids the tedium of manual retention rate design.

\textbf{Comparison with SOTA Backbone Models.} We further compare the model complexity and accuracy of ToFe on LV-ViT with other SOTA backbone models on ImageNet dataset. As shown in Fig.~\ref{FLOPSvsAccMODELS}, our ToFe-based LV-ViT shows competitive performance under different model complexities compared with other CNN-based and ViT-based models, as well as advanced accuracy. Specifically, ToFe compresses half of the FLOPs on LV-ViT-M and LV-ViT-S while maintaining superior accuracy compared to other models, making LV-ViT compressed by ToFe more suitable for widespread deployment on resource-constrained edge environments.

\subsection{Ablation Study}

\begin{table}[t]
  \centering
  \caption{Performance of ToFe with/without token reusing.}
    \begin{tabular}{ccc}
    \toprule
    Model & GFLOPs & TOP-1 Acc. \\
    \midrule
    LV-ViT-S & 6.6   & 83.3 \\
    \midrule
    EViT~\cite{liang2022not}  & 3.4   & 75.1 \\
    DynamicViT~\cite{rao2021dynamicvit} & 3.3   & 77.9 \\
    AS-ViT~\cite{liu2023adaptive} & 3.4   & 63.5 \\
    IdleViT~\cite{xu2023no} & 3.3   & 79.5 \\
    \midrule
    ToFe w/o token reusing & 3.3   & 79.2 \\
    ToFe w/ token reusing & 3.3   & \textbf{81.9} \\
    \bottomrule
    \end{tabular}%
  \label{w/o}%
\end{table}%

\begin{table}[t]
\caption{Performance of different token selector structures.}
    \centering
        \begin{tabular}{cccc}
        \toprule
        Structure & GFLOPs & Keep Ratio & Top1-Acc \\
        \midrule
        3-layer MLP & 2.3 & (0.49,0.25,0.14) & \textbf{78.48} \\
        \midrule
        DynamicViT & 2.3 & (0.47,0.21,0.11) & 74.35 \\
        \bottomrule
        \end{tabular}
    \label{Selector}
\end{table}

\textbf{Impact of Token Reusing.} 
To demonstrate the performance advantages of token reusing, we compared the accuracy of ToFe with/without token reusing on the LV-ViT-S model under identical computational budgets. Similar to DynamicViT~\cite{rao2021dynamicvit} and AS-ViT~\cite{liu2023adaptive}, ToFe without token reusing employs the Hadamard product ${\mathbf{M}^s} \leftarrow {\mathbf{M}^{s - 1}} \odot {\mathbf{M}^s}$ to update the binary decision masks $\mathbf{M}^{s}$, which indicates that insignificant tokens will not be reused in subsequent transformer blocks. As shown in Table.~\ref{w/o} ToFe without token reusing suffers a 2.7\% accuracy decline. Although it still outperforms most SOTA token reduction methods, it falls short of token reusing approaches like ToFe and IDleViT.

\begin{table*}[htbp]
\caption{Performance comparison with different structures of the token approximator.}
    \centering
        \begin{tabular}{ccccc}
        \toprule
        Structure of the token approximator & GFLOPs &  \makecell{\% of GFLOPs caused by \\the token approximator} & Keep Ratio & Top1-Acc \\
        \midrule
        Bottleneck MLP(Ours) & 2.3 & 2.52\% & (0.49,0.25,0.14) & \textbf{78.48} \\
        \midrule
        Identity & 2.3  & - & (0.59,0.26,0.14) & 77.09 \\
        Depth-wise CONV & 2.3 & 0.059\% & (0.55,0.25,0.14) & 77.33 \\
        Transformer Block & 2.4 & 31.67\% & (0.27,0.11,0.07) & 74.27  \\
        \bottomrule
        \end{tabular}
    \label{Approx}
\end{table*}

\textbf{Structure of the Token Selector.} As shown in Fig. \ref{ToFe}, we use a simple but efficient MLP structure for the token selector module in the proposed ToFe framework. However, previous work utilized complex network structures for token selectors. For example, a series of MLP were used in DynamicViT ~\cite{rao2021dynamicvit} for token selectors, where these MLPs worked together with average and concatenation operations and were divided to be responsible for global and local information, respectively. However, our experiment results show that simply using a 3-layer MLP structure is enough for token selectors. In our experiment, we changed ToFe's token selectors to the same structure as DynamicViT while the remaining modules were unchanged. As shown in Table.~\ref{Selector}, changing the structure of the token selector causes more than a 4\% drop in accuracy. At the same time, the keep ratio in each stage decreases under the same computation budget. This result indicates a complex token selector is not necessary because it is difficult to learn due to its complexity. As the token selector brings in additional computation, the actual number of tokens that can engage in inference decreases. The above observations show the 3-layer MLP structure is good enough as an effective token selector.

\begin{table}[t]
\caption{Performance comparison with different keep ratios.}
  \centering
    \begin{tabular}{ccc}
    \toprule
    Keep Ratio & GFLOPs & Top1-Acc \\
    \midrule
    Optimized: (0.49,0.25,0.14) & 2.3   & \textbf{78.48} \\
    \midrule
    (0.5,0.25,0.125) & 2.3   & 78.35 \\
    (0.4,0.4,0.1) & 2.3   & 78.21 \\
    (0.33,0.33,0.33) & 2.3   & 76.78 \\
    (0.4,0.3,0.3) & 2.3   & 77.86 \\
    \bottomrule
    \end{tabular}
  \label{keepratio}
\end{table}

\textbf{Structure of the Token Approximator.} The token approximator is another critical module in ToFe, which requires to balance of high token approximation accuracy and computational efficiency. Table~\ref{Approx} compares the bottleneck structure in ToFe with three alternative implementations.
First, the naive method is to directly reuse the tokens, where the token approximation is treated as an identity function. Second, depth-wise convolution ~\cite{chollet2017xception} is introduced as a lightweight alternative for feature extraction, with behavior resembling local attention. Finally, a transformer block offers a more powerful feature extraction capability but comes with substantial computational overhead.
As shown in Table~\ref{Approx}, while direct token reuse and depth-wise convolution are computationally efficient, they exhibit non-negligible errors and perform worse than the bottleneck structure despite retaining more tokens. Conversely, the high computational overhead of transformer block restricts the number of tokens retained during inference, ultimately also leading to a loss in accuracy. By comparison, the bottleneck structure achieves the best trade-off between model complexity and approximation accuracy, outperforming all other evaluated alternatives.

\textbf{Computation Budget-aware Optimized Keep Ratio.} ToFe optimizes the number of tokens used at each stage within a given computation budget, thereby achieving superior performance. To further validate this, we compare the computation budget-aware optimized keep ratio with fixed keep ratios. We use the DeiT-S model with 50\% FLOPs as the baseline. The experimental results are shown in Table~\ref{keepratio}. The optimized keep ratio is the average value recorded during the training process. It is observed that ToFe with an optimized keep ratio achieves the best performance for the same FLOPs, indicating that ToFe is able to adaptively allocate token usage across different stages according to the computation budget, thereby attaining a better trade-off between model complexity and accuracy.

\begin{table}[t]
    \caption{Performance comparison with different batch size.}
    \centering
    \begin{tabular}{cccccc}
        \toprule
        Batch Size & 1   & 16    & 32    & 64    & 128 \\
        \midrule
        Top-1 Acc (\%) & 78.62 & 78.58 & 78.51 & 78.45 & 78.48 \\
        \midrule
        GFLOPs & 2.3   & 2.3   & 2.3   & 2.3   & 2.3 \\
        \bottomrule
    \end{tabular}
    \label{batchsize}
\end{table}

\textbf{Impact of the Batch Size for Parallel.} As described in Sec.~\ref{optimize}, we propose \texttt{instance-adaptive ToFe} and \texttt{batch-adaptive ToFe}. Among them, \texttt{instance-adaptive ToFe} is well-suited for single-instance inference, while \texttt{batch-adaptive ToFe} is designed for parallel inference. Since we use the average FLOPs of one mini-batch to calculate the computation budget-aware loss during training, parallel inference can be implemented by utilizing the average number of tokens used at each stage as measured during training. The experimental results in Table~\ref{batchsize} demonstrate that batch inference does not cause significant performance degradation.

\subsection{Visualization}

\begin{figure}[t]
\centering
\includegraphics[width=0.48\textwidth]{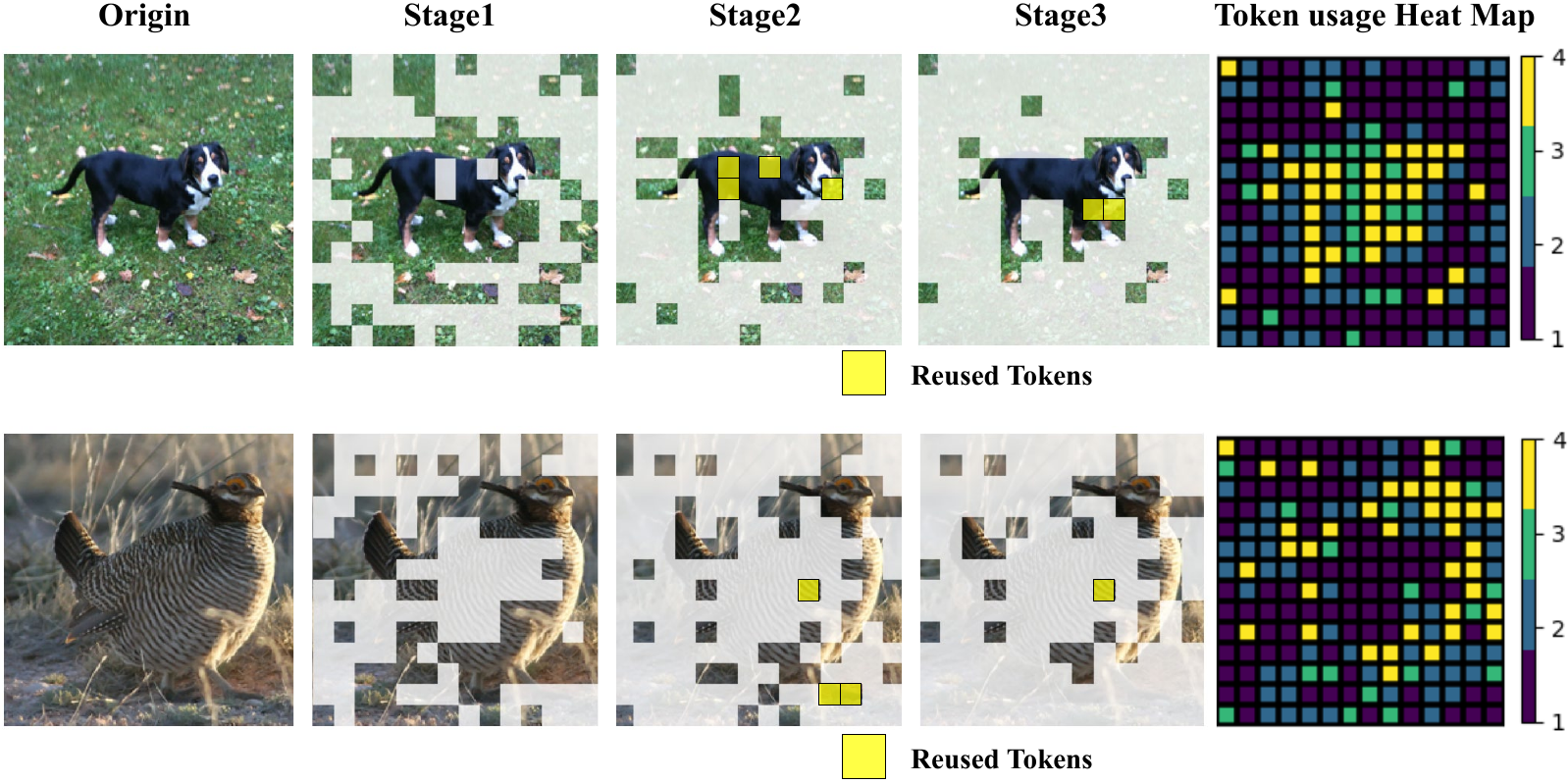}
\caption{Visualization of tokens used at each stage.}
\label{usage_vis}
\end{figure}

\begin{table}[t]
  \centering
  \caption{Statistic of the reused tokens at different stage}
    \begin{tabular}{cccc}
    \toprule
          & \# Remaining Tokens & \# Reused Tokens & Ratio \\
    \midrule
    \multicolumn{4}{c}{ Backbone Model: DeiT-S} \\
    \midrule
    Stage 2 & 48.71 & 5.32  & 10.92\% \\
    Stage 3 & 26.43 & 2.68  & 10.14\% \\
    \midrule
    \multicolumn{4}{c}{ Backbone Model: LV-ViT-S} \\
    \midrule
    Stage 2 & 40.67 & 4.58  & 11.26\% \\
    Stage 3 & 28.03 & 3.26  & 11.63\% \\
    \bottomrule
    \end{tabular}%
  \label{tokens}%
\end{table}%

\textbf{Visualization of Token Freezing and Reusing.} First, we demonstrate the process of token freezing and reusing in ToFe. 
Fig.~\ref{usage_vis} visualizes the token reduction in three stages, where gray squares represent tokens that have been reduced, and yellow squares mark the reused tokens that have been frozen in the earlier stage. Although the number of reused tokens is modest, these tokens are rich in task-relevant information, such as the body of the dog or the feathers and claws of the chicken. These tokens contain features that are critical for image classification. Therefore, even if they are frozen in the shallow blocks, ToFe still can reuse them in later blocks. Additionally, the last column in Fig.~\ref{usage_vis} shows the frequency each token is used. Since ToFe uses a three-stage freezing and reusing framework, each token can be used up to 4 times. The heat map shows that the most informative area in the input images corresponds to the highest token usage, indicating that ToFe effectively preserves the most important tokens, rather than erroneously discarding them. Moreover, Table.~\ref{tokens} shows the statistic results of the number of reused tokens retained in stages 2 and 3 on the ImageNet-1K validation set based on DeiT-S and LV-ViT-S model, respectively. Although the results show that reused tokens constitute a relatively small fraction at each stage, Fig.~\ref{usage_vis} reveals that these tokens contain detailed and important information, indicating it is critical to reuse such tokens in deeper layers.

\begin{figure}[t]
\centerline{\includegraphics[scale=0.45]{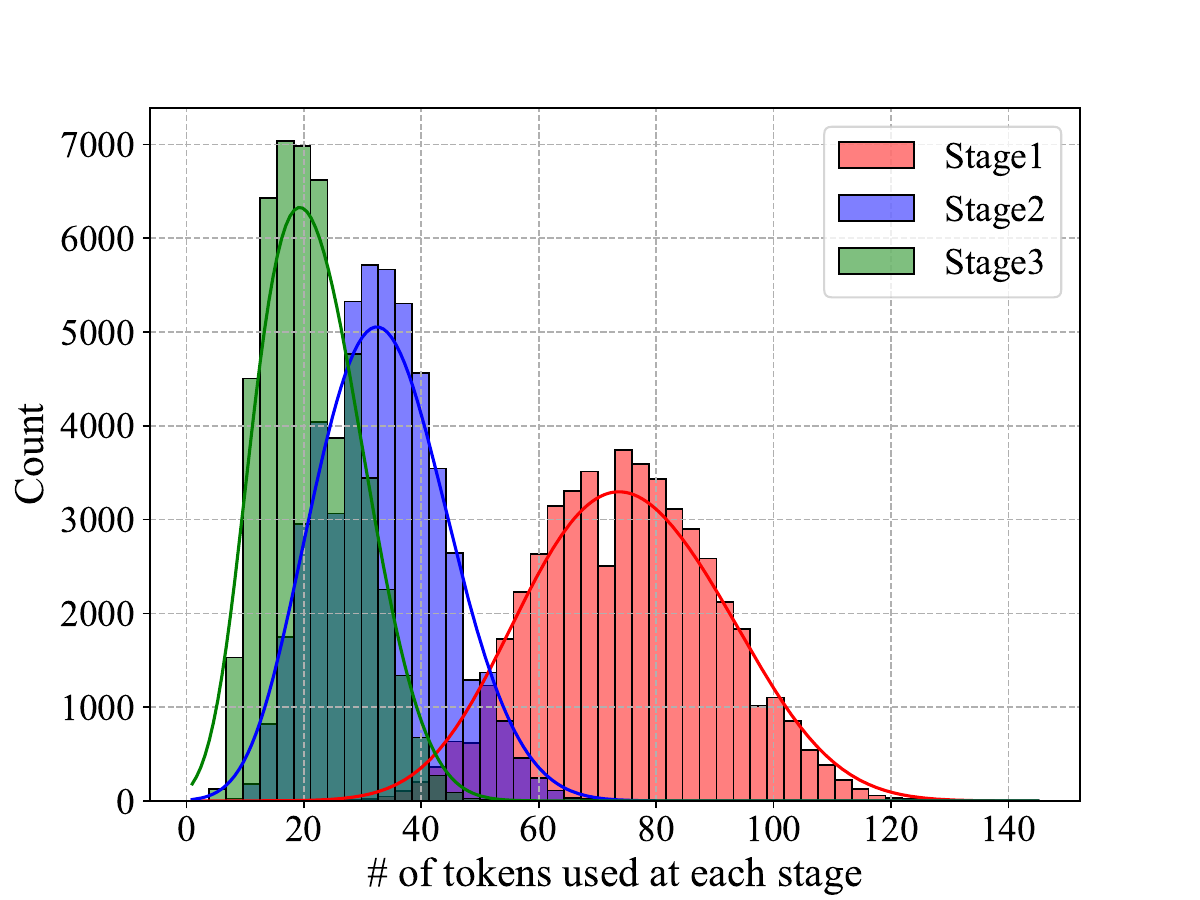}}
\caption{Distribution of the token usage at different stages.}
\label{usage_count}
\end{figure}

\begin{figure}[t]
\centering
\includegraphics[width=0.48\textwidth]{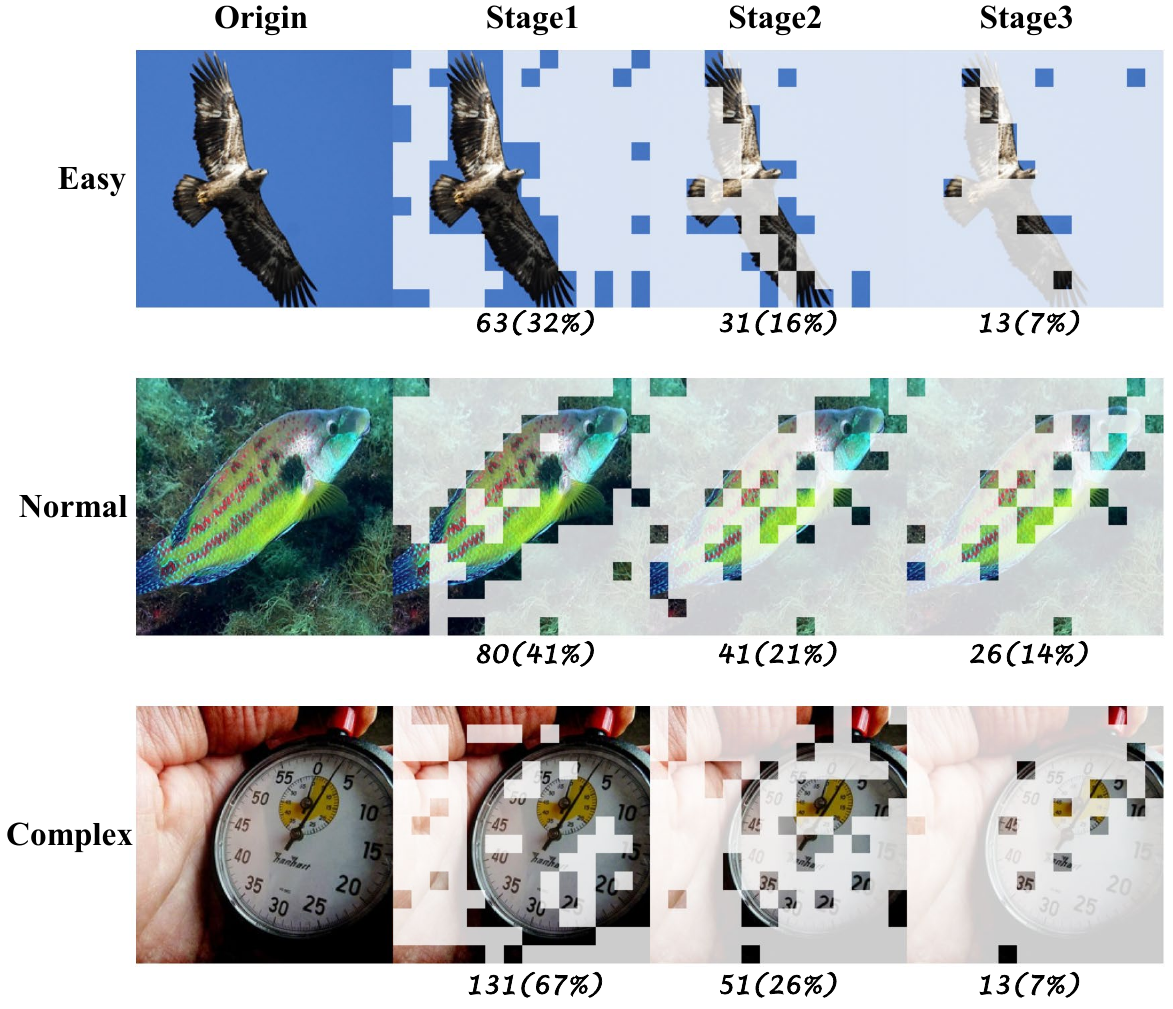}
\caption{Visualization of token usage for samples with different complexities.}
\label{adapvis}
\end{figure}

\textbf{Visualization of Adaptive Token Usage.} Instance-adaptive ToFe can adaptively select the keep ratio at each stage based on the complexity of the input image. To analyze this advantage, we summarize the distribution of the number of tokens used at each stage on ImageNet validation dataset. As shown in Fig.~\ref{usage_count}, the number of tokens used at each stage follows a Gaussian distribution with a specific mean value, indicating that ToFe adaptively makes the corresponding token usage decisions according to the complexity of the input images. Furthermore, we select three images with different complexities to visualize the token usage at each stage. As shown in Fig.~\ref{adapvis}, ToFe can achieve a high token reduction ratio by preserving fewer tokens at early stages to save computational cost. For complicated images, it is difficult to identify the important tokens at earlier stage, thus ToFe tends to retain more tokens at earlier stages while using fewer tokens at later stages. This adaptive mechanism ensures that ToFe strikes a balance between computational efficiency and the retention of crucial information for accurate classification.

\section{Conclusion}

In this paper, we propose a lagged token freezing and reusing framework to implement adaptive usage of tokens. By identifying and temporarily freezing less important tokens, ToFe allows for their reuse in later blocks, ensuring that potentially useful information is not discarded prematurely. Additionally, we adopt a computation budget-aware framework, which enables the global optimization of token selection and flexible token reusing. Extensive experiments demonstrate that ToFe not only reduces the computational cost but also maintains performance, striking a balance between efficiency and accuracy. In the future, we will adapt ToFe to various downstream tasks and combine it with other efficient ViT architectures for further acceleration.




\bibliographystyle{IEEEtran}
\bibliography{IEEEabrv,reference}

\begin{thebibliography}{10}
\providecommand{\url}[1]{#1}
\csname url@samestyle\endcsname
\providecommand{\newblock}{\relax}
\providecommand{\bibinfo}[2]{#2}
\providecommand{\BIBentrySTDinterwordspacing}{\spaceskip=0pt\relax}
\providecommand{\BIBentryALTinterwordstretchfactor}{4}
\providecommand{\BIBentryALTinterwordspacing}{\spaceskip=\fontdimen2\font plus
\BIBentryALTinterwordstretchfactor\fontdimen3\font minus \fontdimen4\font\relax}
\providecommand{\BIBforeignlanguage}[2]{{%
\expandafter\ifx\csname l@#1\endcsname\relax
\typeout{** WARNING: IEEEtran.bst: No hyphenation pattern has been}%
\typeout{** loaded for the language `#1'. Using the pattern for}%
\typeout{** the default language instead.}%
\else
\language=\csname l@#1\endcsname
\fi
#2}}
\providecommand{\BIBdecl}{\relax}
\BIBdecl

\bibitem{vaswani2017attention}
A.~Vaswani, N.~Shazeer, N.~Parmar, J.~Uszkoreit, L.~Jones, A.~N. Gomez, {\L}.~Kaiser, and I.~Polosukhin, ``Attention is all you need,'' \emph{Advances in neural information processing systems}, vol.~30, 2017.

\bibitem{devlin2018bert}
J.~Devlin, M.-W. Chang, K.~Lee, and K.~Toutanova, ``Bert: Pre-training of deep bidirectional transformers for language understanding,'' \emph{arXiv preprint arXiv:1810.04805}, 2018.

\bibitem{chatgpt2022}
\BIBentryALTinterwordspacing
OpenAI. (2022) Introducing chatgpt. [Online]. Available: \url{https://openai.com/index/chatgpt/}
\BIBentrySTDinterwordspacing

\bibitem{copilot2023}
\BIBentryALTinterwordspacing
Github. (2023) Github copilot: Your ai pair programmer. [Online]. Available: \url{https://github.com/features/copilot}
\BIBentrySTDinterwordspacing

\bibitem{touvron2021training}
H.~Touvron, M.~Cord, M.~Douze, F.~Massa, A.~Sablayrolles, and H.~J{\'e}gou, ``Training data-efficient image transformers \& distillation through attention,'' in \emph{International conference on machine learning}.\hskip 1em plus 0.5em minus 0.4em\relax PMLR, 2021, pp. 10\,347--10\,357.

\bibitem{ren2023tinymim}
S.~Ren, F.~Wei, Z.~Zhang, and H.~Hu, ``Tinymim: An empirical study of distilling mim pre-trained models,'' in \emph{Proceedings of the IEEE/CVF Conference on Computer Vision and Pattern Recognition}, 2023, pp. 3687--3697.

\bibitem{ma2023llm}
X.~Ma, G.~Fang, and X.~Wang, ``Llm-pruner: On the structural pruning of large language models,'' \emph{Advances in neural information processing systems}, vol.~36, pp. 21\,702--21\,720, 2023.

\bibitem{yu2022width}
F.~Yu, K.~Huang, M.~Wang, Y.~Cheng, W.~Chu, and L.~Cui, ``Width \& depth pruning for vision transformers,'' in \emph{Proceedings of the AAAI Conference on Artificial Intelligence}, vol.~36, no.~3, 2022, pp. 3143--3151.

\bibitem{ding2022towards}
Y.~Ding, H.~Qin, Q.~Yan, Z.~Chai, J.~Liu, X.~Wei, and X.~Liu, ``Towards accurate post-training quantization for vision transformer,'' in \emph{Proceedings of the 30th ACM international conference on multimedia}, 2022, pp. 5380--5388.

\bibitem{wang2021pyramid}
W.~Wang, E.~Xie, X.~Li, D.-P. Fan, K.~Song, D.~Liang, T.~Lu, P.~Luo, and L.~Shao, ``Pyramid vision transformer: A versatile backbone for dense prediction without convolutions,'' in \emph{Proceedings of the IEEE/CVF international conference on computer vision}, 2021, pp. 568--578.

\bibitem{xu2021co}
W.~Xu, Y.~Xu, T.~Chang, and Z.~Tu, ``Co-scale conv-attentional image transformers,'' in \emph{Proceedings of the IEEE/CVF international conference on computer vision}, 2021, pp. 9981--9990.

\bibitem{liu2021swin}
Z.~Liu, Y.~Lin, Y.~Cao, H.~Hu, Y.~Wei, Z.~Zhang, S.~Lin, and B.~Guo, ``Swin transformer: Hierarchical vision transformer using shifted windows,'' in \emph{Proceedings of the IEEE/CVF international conference on computer vision}, 2021, pp. 10\,012--10\,022.

\bibitem{yuan2021tokens}
L.~Yuan, Y.~Chen, T.~Wang, W.~Yu, Y.~Shi, Z.-H. Jiang, F.~E. Tay, J.~Feng, and S.~Yan, ``Tokens-to-token vit: Training vision transformers from scratch on imagenet,'' in \emph{Proceedings of the IEEE/CVF international conference on computer vision}, 2021, pp. 558--567.

\bibitem{liang2022not}
Y.~Liang, C.~Ge, Z.~Tong, Y.~Song, J.~Wang, and P.~Xie, ``Not all patches are what you need: Expediting vision transformers via token reorganizations,'' \emph{arXiv preprint arXiv:2202.07800}, 2022.

\bibitem{rao2021dynamicvit}
Y.~Rao, W.~Zhao, B.~Liu, J.~Lu, J.~Zhou, and C.-J. Hsieh, ``Dynamicvit: Efficient vision transformers with dynamic token sparsification,'' \emph{Advances in neural information processing systems}, vol.~34, pp. 13\,937--13\,949, 2021.

\bibitem{bolya2022token}
D.~Bolya, C.-Y. Fu, X.~Dai, P.~Zhang, C.~Feichtenhofer, and J.~Hoffman, ``Token merging: Your vit but faster,'' in \emph{The Eleventh International Conference on Learning Representations}, 2023.

\bibitem{liu2023adaptive}
X.~Liu, T.~Wu, and G.~Guo, ``Adaptive sparse vit: towards learnable adaptive token pruning by fully exploiting self-attention,'' in \emph{Proceedings of the Thirty-Second International Joint Conference on Artificial Intelligence}, 2023, pp. 1222--1230.

\bibitem{liusimple}
D.~Liu, M.~Kan, S.~Shan, and C.~Xilin, ``A simple romance between multi-exit vision transformer and token reduction,'' in \emph{The Twelfth International Conference on Learning Representations}, 2024.

\bibitem{zhang2024synergistic}
Y.~Zhang, L.~Wei, and N.~Freris, ``Synergistic patch pruning for vision transformer: Unifying intra-\& inter-layer patch importance,'' in \emph{The Twelfth International Conference on Learning Representations}, 2024.

\bibitem{chen2023diffrate}
M.~Chen, W.~Shao, P.~Xu, M.~Lin, K.~Zhang, F.~Chao, R.~Ji, Y.~Qiao, and P.~Luo, ``Diffrate: Differentiable compression rate for efficient vision transformers,'' in \emph{Proceedings of the IEEE/CVF International Conference on Computer Vision}, 2023, pp. 17\,164--17\,174.

\bibitem{long2023beyond}
S.~Long, Z.~Zhao, J.~Pi, S.~Wang, and J.~Wang, ``Beyond attentive tokens: Incorporating token importance and diversity for efficient vision transformers,'' in \emph{Proceedings of the IEEE/CVF Conference on Computer Vision and Pattern Recognition}, 2023, pp. 10\,334--10\,343.

\bibitem{wei2023joint}
S.~Wei, T.~Ye, S.~Zhang, Y.~Tang, and J.~Liang, ``Joint token pruning and squeezing towards more aggressive compression of vision transformers,'' in \emph{Proceedings of the IEEE/CVF Conference on Computer Vision and Pattern Recognition}, 2023, pp. 2092--2101.

\bibitem{jiang2021token}
Z.~Jiang, Q.~Hou, L.~Yuan, D.~Zhou, X.~Jin, A.~Wang, and J.~Feng, ``Token labeling: Training a 85.5\% top-1 accuracy vision transformer with 56m parameters on imagenet,'' \emph{arXiv preprint arXiv:2104.10858}, vol.~3, no.~6, p.~7, 2021.

\bibitem{visualize2021}
\BIBentryALTinterwordspacing
(2021) Visualizer. [Online]. Available: \url{https://github.com/luo3300612/Visualizer}
\BIBentrySTDinterwordspacing

\bibitem{jang2016categorical}
E.~Jang, S.~Gu, and B.~Poole, ``Categorical reparameterization with gumbel-softmax,'' \emph{arXiv preprint arXiv:1611.01144}, 2016.

\bibitem{maddison2016concrete}
C.~J. Maddison, A.~Mnih, and Y.~W. Teh, ``The concrete distribution: A continuous relaxation of discrete random variables,'' \emph{arXiv preprint arXiv:1611.00712}, 2016.

\bibitem{jiang2021all}
Z.-H. Jiang, Q.~Hou, L.~Yuan, D.~Zhou, Y.~Shi, X.~Jin, A.~Wang, and J.~Feng, ``All tokens matter: Token labeling for training better vision transformers,'' \emph{Advances in neural information processing systems}, vol.~34, pp. 18\,590--18\,602, 2021.

\bibitem{deng2009imagenet}
J.~Deng, W.~Dong, R.~Socher, L.-J. Li, K.~Li, and L.~Fei-Fei, ``Imagenet: A large-scale hierarchical image database,'' in \emph{2009 IEEE conference on computer vision and pattern recognition}.\hskip 1em plus 0.5em minus 0.4em\relax Ieee, 2009, pp. 248--255.

\bibitem{chen2021crossvit}
C.-F.~R. Chen, Q.~Fan, and R.~Panda, ``Crossvit: Cross-attention multi-scale vision transformer for image classification,'' in \emph{Proceedings of the IEEE/CVF international conference on computer vision}, 2021, pp. 357--366.

\bibitem{chu2021conditional}
X.~Chu, Z.~Tian, B.~Zhang, X.~Wang, and C.~Shen, ``Conditional positional encodings for vision transformers,'' \emph{arXiv preprint arXiv:2102.10882}, 2021.

\bibitem{radosavovic2020designing}
I.~Radosavovic, R.~P. Kosaraju, R.~Girshick, K.~He, and P.~Doll{\'a}r, ``Designing network design spaces,'' in \emph{Proceedings of the IEEE/CVF conference on computer vision and pattern recognition}, 2020, pp. 10\,428--10\,436.

\bibitem{tan2019efficientnet}
M.~Tan and Q.~Le, ``Efficientnet: Rethinking model scaling for convolutional neural networks,'' in \emph{International conference on machine learning}.\hskip 1em plus 0.5em minus 0.4em\relax PMLR, 2019, pp. 6105--6114.

\bibitem{brock2021high}
A.~Brock, S.~De, S.~L. Smith, and K.~Simonyan, ``High-performance large-scale image recognition without normalization,'' in \emph{International conference on machine learning}.\hskip 1em plus 0.5em minus 0.4em\relax PMLR, 2021, pp. 1059--1071.

\bibitem{pan2021ia}
B.~Pan, R.~Panda, Y.~Jiang, Z.~Wang, R.~Feris, and A.~Oliva, ``Ia-red2: Interpretability-aware redundancy reduction for vision transformers,'' \emph{Advances in Neural Information Processing Systems}, vol.~34, pp. 24\,898--24\,911, 2021.

\bibitem{xu2022evo}
Y.~Xu, Z.~Zhang, M.~Zhang, K.~Sheng, K.~Li, W.~Dong, L.~Zhang, C.~Xu, and X.~Sun, ``Evo-vit: Slow-fast token evolution for dynamic vision transformer,'' in \emph{Proceedings of the AAAI Conference on Artificial Intelligence}, vol.~36, no.~3, 2022, pp. 2964--2972.

\bibitem{kim2024token}
M.~Kim, S.~Gao, Y.-C. Hsu, Y.~Shen, and H.~Jin, ``Token fusion: Bridging the gap between token pruning and token merging,'' in \emph{Proceedings of the IEEE/CVF Winter Conference on Applications of Computer Vision}, 2024, pp. 1383--1392.

\bibitem{dosovitskiy2021an}
A.~Dosovitskiy, L.~Beyer, A.~Kolesnikov, D.~Weissenborn, X.~Zhai, T.~Unterthiner, M.~Dehghani, M.~Minderer, G.~Heigold, S.~Gelly, J.~Uszkoreit, and N.~Houlsby, ``An image is worth 16x16 words: Transformers for image recognition at scale,'' in \emph{International Conference on Learning Representations}, 2021.

\bibitem{devlin2019bert}
J.~Devlin, M.-W. Chang, K.~Lee, and K.~Toutanova, ``Bert: Pre-training of deep bidirectional transformers for language understanding,'' in \emph{Proceedings of the 2019 Conference of the North American Chapter of the Association for Computational Linguistics: Human Language Technologies, Volume 1 (Long and Short Papers)}, 2019, pp. 4171--4186.

\bibitem{zhang2021edge}
J.~Zhang, Z.~Qu, C.~Chen, H.~Wang, Y.~Zhan, B.~Ye, and S.~Guo, ``Edge learning: The enabling technology for distributed big data analytics in the edge,'' \emph{ACM Computing Surveys (CSUR)}, vol.~54, no.~7, pp. 1--36, 2021.

\bibitem{chen2024otas}
J.~Chen, W.~Xu, Z.~Hong, S.~Guo, H.~Wang, J.~Zhang, and D.~Zeng, ``Otas: An elastic transformer serving system via token adaptation,'' \emph{arXiv preprint arXiv:2401.05031}, 2024.

\bibitem{vig2019analyzing}
J.~Vig and Y.~Belinkov, ``Analyzing the structure of attention in a transformer language model,'' \emph{arXiv preprint arXiv:1906.04284}, 2019.

\bibitem{chollet2017xception}
F.~Chollet, ``Xception: Deep learning with depthwise separable convolutions,'' in \emph{Proceedings of the IEEE conference on computer vision and pattern recognition}, 2017, pp. 1251--1258.

\bibitem{cybenko1989approximation}
G.~Cybenko, ``Approximation by superpositions of a sigmoidal function,'' \emph{Mathematics of control, signals and systems}, vol.~2, no.~4, pp. 303--314, 1989.

\bibitem{bonnaerens2023learned}
M.~Bonnaerens and J.~Dambre, ``Learned thresholds token merging and pruning for vision transformers,'' \emph{arXiv preprint arXiv:2307.10780}, 2023.

\bibitem{wu2023ppt}
X.~Wu, F.~Zeng, X.~Wang, and X.~Chen, ``Ppt: Token pruning and pooling for efficient vision transformers,'' \emph{arXiv preprint arXiv:2310.01812}, 2023.

\bibitem{xu2023no}
X.~Xu, C.~Li, Y.~Chen, X.~Chang, J.~Liu, and S.~Wang, ``No token left behind: Efficient vision transformer via dynamic token idling,'' in \emph{Australasian Joint Conference on Artificial Intelligence}.\hskip 1em plus 0.5em minus 0.4em\relax Springer, 2023, pp. 28--41.

\end{thebibliography}


 





\end{document}